\documentclass{article}

\usepackage{arxiv}

\usepackage[utf8]{inputenc} % allow utf-8 input
\usepackage[T1]{fontenc}    % use 8-bit T1 fonts
\usepackage{hyperref}       % hyperlinks
\usepackage{url}            % simple URL typesetting
\usepackage{booktabs}       % professional-quality tables
\usepackage{amsfonts}       % blackboard math symbols
\usepackage{nicefrac}       % compact symbols for 1/2, etc.
\usepackage{microtype}      % microtypography
\usepackage{lipsum}		% Can be removed after putting your text content
\usepackage{graphicx}
\usepackage{natbib}
\usepackage{doi}

\usepackage{xcolor}
\usepackage{xspace}
\usepackage{subfig}
\usepackage{graphicx}
\usepackage{fixmath}

\usepackage{wasysym}
\newcommand\doublecheck{\textcolor{black}{\checked\kern-0.6em\checked}}

\input{glyphtounicode}  % With this ``fi'' and other ligatures are copied correctly from the PDF
\usepackage{array}
\PassOptionsToPackage{hyphens}{url}
\usepackage{hyperref}

\usepackage{xspace}	% in flach
\usepackage{pgf}  % colours

\usepackage{floatrow}
\usepackage{subfig}

\usepackage{amsopn}

\usepackage{color}
\definecolor{darkred}{rgb}{0.55, 0.0, 0.0}
\definecolor{orange}{RGB}{255,127,0}
\definecolor{brown}{RGB}{150,70,0}
\definecolor{green}{RGB}{127,255,127}
\definecolor{darkgreen}{RGB}{0,127,0}
\definecolor{blue}{RGB}{127,127,255}
\definecolor{lightblue}{RGB}{150,150,255}
\definecolor{darkblue}{RGB}{0,0,127}
\definecolor{red}{RGB}{255,90,90}
\definecolor{grey}{RGB}{127,127,127}
\definecolor{pink}{RGB}{255,180,180}

\usepackage{xcolor}
\usepackage{nicefrac}
\usepackage{multicol}
\usepackage{booktabs}

\usepackage[normalem]{ulem}
\usepackage{threeparttable}

\newcommand{\inserted}[1]{\textcolor{black}{{#1}}}
\newcommand{\modified}[1]{\textcolor{black}{{#1}}}

\newcommand{\insertedTIST}[1]{\textcolor{black}{{#1}}}
\newcommand{\modifiedTIST}[1]{\textcolor{black}{{#1}}}

\newcommand{\moved}[1]{\textcolor{black}{{#1}}}

\newcommand{\Utility}{\ensuremath{U}\xspace}
\newcommand{\utility}{\ensuremath{u}\xspace}

\newcommand{\Benefit}{\ensuremath{B}\xspace}
\newcommand{\Cost}{\ensuremath{C}\xspace}

\newcommand{\systemCost}{\ensuremath{\mbox{\textsf{\emph{C}}}}\xspace}

\newcommand{\applicationCost}{\ensuremath{\mbox{\textsf{\cal{C}}}}\xspace}

\newcommand{\applicationBenefit}{\ensuremath{{\cal{B}}\xspace}}

\newcommand{\vect}[1]{\ensuremath{\bar{#1}}\xspace}

\title{Between Progress and Potential Impact of AI: \\the Neglected Dimensions}

\author{ \href{https://orcid.org/0000-0003-2902-6477
}{\includegraphics[scale=0.06]{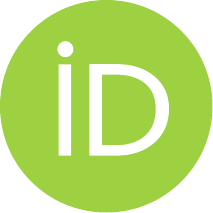}\hspace{1mm}Fernando Mart\'inez-Plumed}\thanks{Corresponding author} \\
	VRAIN, Universitat Polit\`ecnica de Val\`encia\\
	\texttt{fmartinez@dsic.upv.es} \\
	\And
	{Shahar Avin} \\
	CSER, University of Cambridge\\
	\texttt{sa478@cam.ac.uk} \\
	\And
	{Miles Brundage} \\
	OpenAI\\
	\texttt{miles.brundage@asu.edu} \\
	\AND
	{Allan Dafoe}\\
    DeepMind\\
    \texttt{allandafoe@deepmind.com}
	\And
	{Se\'an \'O h\'Eigeartaigh} \\
	CSER, University of Cambridge\\
	\texttt{so348@cam.ac.uk} \\
	\And
	{Jos\'e Hern\'andez-Orallo}\\
    VRAIN, Universitat Polit\`ecnica de Val\`encia\\
    \texttt{jorallo@dsic.upv.es}
}

\renewcommand{\shorttitle}{Between Progress and Potential Impact of AI: the Neglected Dimensions}

\hypersetup{
pdftitle={Between Progress and Potential Impact of AI: the Neglected Dimensions},
}

\begin{document}
\maketitle

\begin{abstract}
	We reframe the analysis of progress in AI by incorporating into an overall framework both the task performance of a system, and the time and resource costs incurred in the development and deployment of the system. These costs include: data, expert knowledge, human oversight, software resources, computing cycles, hardware and network facilities, and (what kind of) time. These costs are distributed over the life cycle of the system, and may place differing demands on different developers and users. The multidimensional performance and cost space we present can be collapsed to a single utility metric that measures the value of the system for different stakeholders. Even without a single utility function, AI advances can be generically assessed by whether they expand the Pareto surface. We label these types of costs as neglected dimensions of AI progress, and explore them using four case studies: Alpha* (Go, Chess, and other board games), ALE (Atari games), ImageNet (Image classification) and Virtual Personal Assistants (Siri, Alexa, Cortana, and Google Assistant). This broader model of progress in AI will lead to novel ways of estimating the potential societal use and impact of an AI system, and the establishment of milestones for future progress.
\end{abstract}

% keywords can be removed

\section{Introduction}

\inserted{The impact a new technology has on society depends on many factors and interests, but it is undeniably and ultimately linked to how powerful and versatile the technology is. This is a two-way association, as measuring the actual progress of a technology depends on how transformative it is for society. AI is no different. In this paper we argue that we need a more thorough account of the elements that (1) play a major role on how efficient it is to deploy a new AI technology or (2) imply some unaccounted costs on the AI life cycle or on society as a whole.}
%
%\removed{Metrics of scientific progress can play an outsized role in the perception of a field and in the allocation of its resources. By contrast, that which goes unmeasured is often neglected. %often goes unnoticed. We argue for a more general accounting of progress in AI, so as to better map attention and metrics to scientific achievement.}

%Policy-makers and research funding bodies, however, are especially interested in the assessment and forecasting of AI progress, and are particularly eager to ask AI researchers for comprehensive criteria and models to evaluate ideas, techniques and products in AI. For instance, when asked whether more money should be invested in developing hardware or algorithms, the answer may change depending on the expert’s domain. 

The prevailing approach to assessing 
%form of objective evidence about 
AI progress consists of measuring \textit{performance}, such as the raw or normalized score in a game, ELO rating, error rate, accuracy, and so forth. 
%, \inserted{or even using meta-measures to compare the evaluation measures themselves, such as calibration measures \cite{ferri2009experimental} or precision-recall for objects and parts in image segmentation \cite{pont2013measures}} \todo{Add a comment about possible meta-measures, but in the end also about performance: \cite{pont2013measures,ferri2009experimental}}
\modified{All these measures are often plotted over time to evaluate temporal progress} \cite{eff2017,eckersley2017measuring}.  
%, the reports usually include  tables and plots with the performance of a range of techniques for some popular tasks (e.g., voice recognition, face detection, game playing, etc.), showing  significant progress over time \cite{eff2017,eckersley2017measuring}. 
%There are several reasons for this choice. First, it is difficult to explain some technical breakthroughs without showing the effect they have on tasks. Second, tasks are usually associated with potential applications. Third, AI is usually conflated with automation, and this ultimately refers to tasks. Fourth, performance is a tangible and quantifiable goal. Fifth, performance is agnostic to how the result is obtained. What could be better than optimal performance?
%
Performance, however, does not exactly correspond with social value or scientific progress in AI. Misalignment between what is measured and what is desired can lead to misallocation of energy and resources. Specifically, excessive effort is likely to go towards achieving novel performance \modified{milestones} \cite{martinez2021research}, 
% benchmarks, %Jose. One reviewer criticised this use of benchmark
and insufficient effort towards progress on other dimensions relevant to social value, economic value, and scientific progress, such as compute efficiency, data efficiency, novelty, replicability, autonomy, and generality. 
%a high degree of automation, compute efficiency, economy, autonomy or, simply, general intelligence, for a wide range of applications.  
%AD to all: I revised the list of other dimensions above. 
%Focusing on performance alone promotes a culture of big undertakings where everything is valid in order to achieve a goal (e.g., win a competition, reach human level, etc.). This task performance race usually fails to identify the ideas, algorithms and systems with a high degree of automation, efficiency, economy, autonomy or, simply, general intelligence, for a wide range of applications. 
%The focus on performance is common in many other disciplines and has been prevalent in AI since at least the first AI winter. However, the culture in AI today, fuelled by investors and conferences looking for breakthroughs, and the access of a few giants to a large amount and variety of resources, is altering the balance.

This does not mean that quantitative assessment
\modified{through benchmarking} %the measurement of quality should be relinquished or benchmarks in AI 
should be abandoned \cite{martinez2016ai,martinez2016making,martinez2018dual,martinez2019item,hernandez2022training,martinez2022ai}. On the contrary, we need more and better measurement \cite{moam2017,hernandez2021general}: measurement which is more comprehensive, general, %less short-term,
%AD to all: I didn't understand the "short-term" reference. 
and focused on the cost function of the ultimate beneficiaries. Ultimately %, in assessing progress 
we would like to weight all the resources that users (or receivers) of a technology require to achieve their goals. For instance, to what extent does progress on a particular metric of performance in machine translation map on to user's satisfaction? Does the progress also correspond to a reduction in cost per translation, or in time for execution? If a paper develops a new technique, how easily can \modified{this be brought from the laboratory to a generally impactful application?} %new algorithms and applications integrate and benefit from it? 

%In general, in assessing progress in AI one must account for the costs of developing and deploying a system, as well as its performance. 
In general, users seek the benefits of high performance (at a set of tasks), while they seek to minimize the costs of developing and deploying their system. Sensitivity to costs is true for individual consumers, firms and developers, as well as other scientists. 
%In this analysis there must always be a balance between performance and cost. %The key point is that the analysis of costs must ultimately be focused on the whole process and the final beneficiary, not on the first conception or evaluation. 
Some kinds of hidden costs can appear during development, when an application is produced, when %it has to be 
reproduced at a large scale, or when 
%it is 
adapted to other domains. Some future costs will be \modified{borne} by future developers or scientists, sometimes referred to as “technical debt” or “research debt”.
%) are internal to the product: they will appear sooner or later and someone will have to cover these costs, from companies to final users. 
%We will see some of these internal hidden costs in AI.
Other costs may be spread more broadly, and are thus harder to account for. 
%can be more difficult to bring to the surface: external costs. 
%\sidenoteAllan{I removed the internal costs and external costs distinction. It didn't seem  precise enough to me; I used other language to express our point. For one, technical debt is a form of externality, so in that sense it is an external cost.}
As in other sectors, there are externalities from AI development and deployment which are important to be aware of; among the negative externalities are %carbon and
environmental footprints, user %intrusion and
privacy, skill atrophy (e.g., the Google effect), opacity in decision making, etc. 
%\sidenoteMiles{I'm not sure what technological atrophy means here.}\sidenoteJose{I added an example, the google effect}\sidenoteMiles{I like the example, though am still somewhat bothered by the phrase "technological atrophy" which (based on Googling ;-) ) doesn't seem to be widely used, and suggests the technology rather than the skill is what's atrophying. Maybe "skill atrophy"?}
Attention to, and ideally measurement of \modified{the possible impact of these side effects} is beneficial, as it is a first step towards internalizing them.%\sidenoteAllan{Jose: what do you think of the above? I revised, trying to preserve the spirit of what we were saying.} 
%is warranted, since  Ideally the external costs (and benefits) will be measured and internalized by developers.  
%We are not only talking about job displacement, whose impact will be covered by the state or the whole society, but costs related to energy consumption (and their carbon footprint), more needed online communication (and their network footprint), more hardware needed (and their environmental footprint), more personalized data (and their intrusion footprint), etc. 
%This ultimate AI footprint is also usually overlooked. %\sidenoteAllan{This section might alienate a reader. It comes across a bit as the beginning of an environmental lecture about AI. Given that this is not the focus of the paper, and these externalities are especially tricky to measure, I suggest toning this down or removing it.}\sidenoteJose{Now it has been reduced to just a short sentence. I like this to be included, even if it is not a main part of the paper anymore.}
%\sidenoteJose{All this looks great to me! I just removed two words to save a line :-)}

In this paper we consider this wide range of costs.
%\comment{, though %, from performance to footprints,
%AD to all: "performance" is not a "cost", "footprints" remains poorly defined. 
%but the paper will 
%narrow the focus of the analysis to
%we focus on the costs \modified{borne} by future developers, such as the costs in computation, data, %knowledge, software, human attention, %manipulation,
%and calendar time.} %, and other dimensions. % often overlooked at several stages of the development of an AI technology.
 We will identify 
%the dimensions that cover the internal resources, 
%should be considered by comprehensive measures of quality and progress, 
%and
how costs are %included
distributed 
%\sidenoteAllan{The previous word here was "included", but it wasn't clear. I used "distributed"; is this what we mean?}\sidenoteJose{Yes!}
depending on the stage in which they are incurred, the number of times they are replicated, and the actor covering each cost. %We are aware that the further we go away from the raw notion of performance the more difficulties arise. Actually, on the most comprehensive end we experience the difficulties of the evaluation of value in economics or even significance in philosophy of science. But 
%We will see that 
%there is still margin to bring all
These dimensions should be 
%brought to the technical arena, and make them 
%made integral elements of 
integral to the measurement of AI progress, \inserted{even if their measurement is not always straightforward}. %\sidenoteJose{This must be emphasized if we don't want the paper to be rejected for being out-of-scope.}. %As a way of approximating these costs, we will build on the notions of reproducibility and replicability, and related notions such as reusability and generality.
As an illustration of these difficulties and how they can be overcome, %at least partially, 
we will analyze %a series of
several case studies where we evaluate performance alongside these other dimensions. As a result, we overhaul the notion of progress in these domains.%\sidenoteAllan{I substantially revised the above. Please check.}\sidenoteJose{Fine with me. Only three words in the last paragraph, if someone comes up with a way of reducing the paragraph}\sidenoteAllan{Revised further, tightening.}

Our paper makes several contributions. First, we offer the most %formal 
detailed and formal analysis to date of the dimensions of AI progress. %, and the first formal one. 
While previous work has attempted to  %formally quantify the intelligence (or more mundanely, the performance) 
quantify progress in the performance  
of a specific system, we more fully account for the resources required and the generality of solutions. 
%ask a distinct question: 
%given information of  %performance and 
%the resources required to achieve that performance and the operating areas of the potential receivers, what can we infer about the progress that the system represents? 
Second, in so doing we surface neglected dimensions of AI progress that %, when better articulated and understood, 
may be optimized more directly. 
%(e.g. we carefully distinguish between replicability and reproducibility). 
Third, we offer a novel framing under Pareto optimality 
%\sidenoteMiles{Pareto optimality framing, or Pareto optimality-oriented framing, or Pareto front-oriented framing? just Pareto seems strange to me for some reason}
% framing 
for assessing performance and costs of an AI system, which suggests a more principled approach to forecasting \inserted{the impact of} future developments in AI, with myriad applications for policy, ethical, and economic analysis, and better research portfolio optimization within the field of AI itself.

%The rest of the paper is organized as follows. The following section gives an account of related work around four issues:  representativeness, specialization, reproducibility and resources. Section 3 identifies the dimensions, the space and their integration. Section 4  \todo{Approach to measure them, case studies, guidelines}

%%%%%%%%%%%%%%%%%%%%%%%%%%%%%%%%%%%%%%%%%%%%%%%%%%%%%%%%%%%%%%%%%%%%%%%%%%%%%%%%%%%%%%%%%%
%%%%%%%%%%%%%%%%%%%%%%%%%%%%%%%%%%%%%%%%%%%%%%%%%%%%%%%%%%%%%%%%%%%%%%%%%%%%%%%%%%%%%%%%%%
\section{Background}\label{sec:back}
Many benchmarks and competitions are used %for all subdomains of
in AI, but they vary in how representative they are of the fundamental problems in their respective subfields \cite{aievaluation2017,aicosmos2017}. 
%\comment{For instance, it has recently been recognized %\citeanom{Morgenstern-personal-communication} 
%that the Winograd Schema challenge only partially represents commonsense reasoning.} 
%\sidenoteJose{Leora Morgenstern, one of the authors and maintainers of the challenge, at the panel of commonsensereasoning.org in 2017 said that. We can't refer to this now without losing anonymity.}. 
%
%In general, this has happened whenever some other tasks that were considered representative of domains of AI have been solved (the “AI effect”). Partly because of this, there is also a strong argument in favor of solving real problems instead of collections of benchmarks. 
As a reaction, challenges in AI are realigned to \modified{see if they can better capture the potential impact on automation}  %achieve more and better automation
\cite{frey2017future,brynjolfsson2017can,aghion2017artificial,korinek2017artificial}, or the aspiration of 
%, from looking towards
more human-like AI \cite{%mikolov2015roadmap,
lake2017building,marcus2018deep}. 
%\sidenoteAllan{This sentence is unclear and needs to be revised.}\sidenoteJose{Hope it is now better phrased.} 
%\sidenoteMiles{I'm not sure exactly what "the actual automation needs" means here}
%\sidenoteJose{Hope it is now better phrased.}
%The choice of those challenges that are really representative of what AI and ML should address is a recurrent topic for CHALEARN \cite{chalearn2018}, an organization that has been devising and running dozens of challenges in AI/ML. 
%In the end, every discipline needs to ask itself the question of what part of it “matters” 
%, but in the end there must be some targets and some way to evaluate whether these targets are met 
%\cite{wagstaff2012machine}. 
%
A deeper concern is that most benchmarks %performance is not a sufficient metric (is not representative of)
are not really fostering  %representative of 
the basic scientific advances needed to move the field forward, be they theoretical advances, explanatory insights, or tools to facilitate other work. This issue of {\em non-representativeness} is partly addressed through the review process, and requirements such as controlling the percentage of papers in different areas \cite{shah2017design}. 
%\sidenoteAllan{Jose: is the above an adequate restatement?}\sidenoteJose{Rewritten a bit. With representativeness here we were talking about the problems/challenges themselves, not the performance metric.}\sidenoteMiles{I'm fine with this.} 
%But the major concern about representativeness, or simply the use of benchmarks, is that many advancements are at a theoretical level, are explanatory or are instrumental pieces for others to get to systems that actually perform. This is of course covered by the criteria at major 
%scientific venues 
%conferences and journals (technical quality, clarity and presentation, novelty/originality, and potential impact or usefulness), especially the last one. However, reviewers have to balance that with quality, which, for many papers, refers to the results for some experiments, on some benchmarks, which must also be judged representative or not for the discipline (between highly relevant to the progress of the field to out of scope). All venues have topics or subject areas,
% and some control that the percentage of accepted papers is similar across areas \cite{shah2017design}\sidenoteJose{Important to mention reviewing processes, etc., otherwise we can have some reviewers thinking that we're ignoring their work. And this paper is mostly targeted to them.}, but the short-term and, especially, long-term effect of better experimental results is still unassessed. %analysing how many of these areas depend on better experimental results on benchmarks and their impact some years later might be enlightening. 

A second issue, \textit{specialization}, is related to representativeness. When a benchmark or competition becomes the target, researchers will have incentives to overly specialize their systems to performance on that benchmark, at the cost of other features of their system, such as generalizability.  
%the efforts are focused on systems that are good at whatever cost, including not being good at any other problem. If the same benchmark is kept equal for a few years, many new techniques will be developed for the sole purpose of getting good performance at the benchmark, but may generalize poorly or be completely useless for other problems. %Generality is sometimes seen as opposed to performance: ``It’s natural to ask: which research results will generalize to new situations? One kind of result which probably won't generalize is: 'algorithm A works better than algorithm B'.'' \cite{grosse2014}.\sidenoteAllan{I don't understand this.}\sidenoteJose{It's a very blunt quote by Grosse that any statement made about an algorithm having better performance than another will not generalize to other situation.}
%
%Generality can refer to ideas, techniques, algorithms or scientific papers, but it can also apply to AI systems, or agents. In this case, the terms ``general-purpose'' or “artificial general intelligence” are used. However, it 
If we had a satisfactory metric of generality then we could use that as a benchmark \inserted{measure}, but it remains an open question how best to operationalize generality
%It is still an open question how to characterize and evaluate a system that shows performance on a wide range of tasks 
\cite{moam2017}, 
%as we have to be precise on this distribution of tasks, 
balancing between putting all the distribution mass \inserted{possibly falling} on a few tasks \cite{legg2007universal}---and not really being general---or distributing it in a block-uniform way---facing the no free lunch theorems \cite{wolpert2012no}.

A third issue is \textit{reproducibility}, and the wider notion of replicability. In AI this was usually understood as requiring the sharing of data and code, % in a matter that permits the reproduction of results, 
but the concept is becoming richer %and more nuanced 
\cite{drummond2009replicability,bonsignorio2015toward,henderson2017deep}. Indeed, we must distinguish between specifically reproducing the results, and replicating the findings with some variations \cite{zwaan2017making}. % (for the rest of the paper we will follow this terminology, despite some others in AI \cite{drummond2009replicability} using the opposite terminology). 
%
%Even if reproducibility and replicability may seem welcome, there is some criticism that just getting the same results (full specific reproducibility) might have a negative impact on the representative and generality of a field. 
Several initiatives have been proposed to facilitate (or even require) 
a wider replicability. % as part of the procedure.
 For instance, with the ''open leaderboards´´ \cite{spohrer2017}, participants have to upload their code so that other participants can make modifications %on them 
and submit another proposal. %Aiming at a more reusable and general replicability, we find the notion of reframing \citeanom{reframing2016}, where a model is not retrained but adapted to a new situation.\sidenoteAllan{I don't understand this last sentence. Perhaps just delete it?} 

Finally, users are generally sensitive to the %resource cost
\modified{effort} of developing and deploying an AI system, which performance benchmarks rarely %explicitly 
take into account. Much AI progress is %said to be attributable 
\modified{attributed} to advances in computational power \cite{reagen2017deep,hwang2018computational}. 
%\sidenoteAllan{The following needs to be revised, as it is unclear what is being argued. As I did not understand it, I held off on doing so lest I distort the argument.}\sidenoteJose{Rephrased.}\sidenoteAllan{Ok. Clear enough.} 
However, it is not straightforward to quantify what exactly can be attributed to software progress,  hardware progress %(which might have been underestimated 
%\cite{reagen2017deep}) and
or several other resources 
%(data, ``human input'', fast networks, etc.) %\citeanom{brundage2016modeling}. Some previous work has tried to determine, for some areas, which part of the progress is due to algorithmic improvement or more computation \cite{grace2013algorithmic,grace2017}. 
\cite{brundage2016modeling,grace2017}. 
%
%Actually, the speedup given by hardware and computer architectures 
%, and its potential, might have been underestimated, especially in deep learning 
%might have been underestimated 
%\cite{reagen2017deep}. %, also because some intermediate calculations are now approximated. 
%Some optimizations approximate the some intermediate  calculations but do not have a significant effect on the accuracy of the results (the outputs are  reproduced, but not the process). 
Accordingly, perhaps it is % easier, and more effective, 
more effective 
to just measure the so-called ``end-to-end performance'', including computational time and quality of the models, such as the recent \emph{DAWNBench} \cite{coleman2017} for deep learning, \insertedTIST{or MLPerf \cite{MLPerf2018} for a variety of AI models and (hardware) chips }.  %, and the associated competition. %, evaluating ``time/cost required to achieve a state-of-the-art accuracy level for common deep learning tasks, as well as inference latency/cost at this state-of-the-art accuracy level''.
Other resources, such as data, are at least as important, especially in % machine learning. 
%It is striking that some systems in the 1990s (the knowledge-based systems or expert systems) were criticized because they needed too much knowledge, while the ``new AI'' today is criticized because it needs too much data. Actually, t
%There is an important effort in making machine learning more data efficient
machine learning\footnote{See \url{https://sites.google.com/site/dataefficientml/bibliography} for a bibliography on data-efficient ML.}. But it seems subjective to determine what input is seen positively or negatively, or even considered as cheating: too much data \modified{(versus better algorithms)}, %supervised or unsupervised)
too much knowledge (constraints, rules or bias), enriched input \cite{bougie2017deep}, etc. %As we will see, t
The question depends mostly on the cost of the resource. %There is nothing wrong with using extra information, provided that information is cheap to obtain, easy to handle, and ultimately available for the real applications. This can be seen as considering humans in the loop, also when referring to resources and costs. 
Human resources (``human computation'') are also 
%typically used 
common 
in AI to increase performance or generality (but at the cost of autonomy). 
%Typically, in AI, when human resources are used during operation (to increase performance or generality, but at the cost of autonomy), we use the term ``human computation''. In this context, the goal is to make a more efficient use of humans. %, even through gradient descent \cite{yang2017mastering}.

Overall, there are many resources involved but, at the moment, there is no integrated framework %or model 
taking into account all of them. Related approaches involve the ideas of utility functions, Pareto-optimal analysis and, most especially, cost-sensitive learning \cite{elkan2001foundations}. %, where decisions are evaluated by comprehensive utility or cost functions.
\citeauthor{turney2002types} (\citeyear{turney2002types}) identifies costs related to inputs and outputs in classification (errors, instability, attributes, labeling, actioning) % unwanted achievement),
data (cases), computation and human preprocessing.  In this paper, we offer a general statement of this idea, applied to AI progress.

In the end, when assessing AI progress in a comprehensive way, one should consider the whole life cycle of research, innovation, production, and reproduction. 
%(or massive application of an AI idea or product). 
Notions such as technical or research debt are becoming more recognized, as they incorporate some costs that are not perceived at early stages of the process 
%(when a paper is written) 
but \modified{have an impact} later on, when the technology or product is put into practice \cite{sculley2015hidden,henderson2017deep,olah2017,desislavov2021compute}. 
%\comment{
%Still, the costs may go beyond the product life cycle and have a global effect, being assumed %(externalized) by the whole society, 
%%. Also, some other parts of the technological or, physical, ecosystems are usually overlooked, 
%referred to as ``footprints'' 
%%, such as ``ecological footprint'', ``application footprint'' or ``memory footprint'' 
%\cite{adamski2018distributed}.
%}

%%%%%%%%%%%%%%%%%%%%%%%%%%%%%%%%%%%%%%%%%%%%%%%%%%%%%%%%%%%%%%%%%%%%%%%%%%%%%%%%%%%%%%%%%%
%%%%%%%%%%%%%%%%%%%%%%%%%%%%%%%%%%%%%%%%%%%%%%%%%%%%%%%%%%%%%%%%%%%%%%%%%%%%%%%%%%%%%%%%%%
\section{Components and integration}\label{sec:comp}
%%%%%%%%%%%%%%%%%%%%%%%%%%%%%%%%%%%%%%%%%%%%%%%%%%%%%%%%%%%%%%%%%%%%%%%%%%%%%%%%%%%%%%%%%%
%%%%%%%%%%%%%%%%%%%%%%%%%%%%%%%%%%%%%%%%%%%%%%%%%%%%%%%%%%%%%%%%%%%%%%%%%%%%%%%%%%%%%%%%%%
We now flesh out a comprehensive list of dimensions that are required for an ``AI system'' to work. We use the term ``system'' in a flexible way, including an agent, an algorithm, a product, etc., proposed in a research paper or by a company. %The term ``AI'' in ``AI system'' is much more cumbersome but needs to be clarified before we analyse any dimension. If we considered systems that have some non-trivial degree of intelligence, we would exclude almost all AI systems today. If we considered systems that “do things that would require intelligence if done by [humans]” (Marvin Minsky), we would also exclude many AI systems today. Trying to be pragmatic here, we could just include all (partially) cognitive tasks that humans can do, from simple arithmetics to taking care of your garden. Of course some of these tasks do not require intelligence, but AI would end up having moving targets (``the AI effect''), which is not unusual in technology (e.g., ``Big Data''). The key thing here seems to be the automation of all tasks humans can do. While most (if not all) purely physical tasks have been automated, the focus of AI would be on the rest. This is perhaps the bias behind some AI research: automate something at whatever cost. \sidenoteAllan{This paragraph seems especially in need of tightening.}

\modifiedTIST{Given the fuzzy contours of AI \cite{ijcaiContours}, one relevant way of assessing the impact of AI technology is through 
%
%human automation is usually recognized as a goal for AI \inserted{(while others are understanding natural intelligence or creating different kinds of intelligence)}.\todo{Probably rephrase sentence} 
%
%However, it is actually difficult to distinguish when reports and forecasts about 
the potential for} ``automation'' \cite{frey2017future,brynjolfsson2017can,aghion2017artificial,korinek2017artificial,martinez2020does,tolan2021measuring}. \modified{However, some of these studies are usually} 
assuming conditions such as ``at a reasonable cost'', ``to a high degree of automation'', etc., versus ``full automation at whatever cost''. The 
%rankings and times of the tasks that one expects to be automated (or not) 
estimated probability of automation for a given task 
might change completely depending on these conditions. 
%
%In practice, if a system is semi-automated, and the part that humans has to do is not very expensive, the combination can be really successful. For instance, a self-driving car that would require assistance from a human driver 1\% of the time could still be a very good contribution. 
In the end, automation is important, but it is the efficiency of the whole system what matters to \inserted{assess its potential impact}, including any ``human computation'' involved. 
%This has the same rationale as in other technologies: producing electricity from the sun is not the ultimate goal (this was done many decades ago), what matters is to do it efficiently (and economically competitive with the costs of the other options). 
This view of efficiency links us directly to the resources involved in an AI system. % and their associated costs.

% We must consider all the resources that are involved in providing the service. 
% %The only bound will be about the periods or stages that are considered (e.g., from conception to a prototype, or from an existing service to its mass production). That means that, in many cases, we may be able to include or exclude the research, technology, knowledge, etc., that led to that solution or to its adaptation or adoption, according to
% How to bound the window or period we are interested in will be analyzed in the next section. At this point, let us introduce a list of resources that frequently appear (more or less explicitly) in AI. This is shown in Table~\ref{tab:resources}.

Table~\ref{tab:resources}
shows 
the resources we identified as frequently involved in developing and deploying AI systems. 
%\sidenoteMiles{should this be caveated in some way, e.g. resourced identified in our analysis or something, such that we're not claiming it's the ultimate list?}\sidenoteJose{Done}\sidenoteMiles{looks good, though I suggest "that are" instead of "to be" in this sentence (change made for now)}
%\noindent 
These resources 
%enumerated in Table~\ref{tab:resources} 
have fuzzy boundaries and are often fungible with each other. For instance, the distinction between data and knowledge is not always clear, and hardware and software may be highly intertwined. Human resources are typically considered under ``manipulation'', but can appear in other resources (e.g., labeled data and teaching a robot %are similar but
might be assigned to $r_d$ and $r_m$ respectively). \inserted{Similarly, $r_t$ represents calendar time, which cannot be accelerated by putting more human resources, as we have to wait for some events to happen---unless we use  simulations or historical data from other domains.} \modified{The existence of these fuzzy boundaries} is not a problem, as long as all the resources are identified. %From the table above we exclude some other resources that are independent of the technology, such as finance and insurance, and cannot be included in other categories (e.g., testing resources involve data, computation, manipulation, etc.). 

\newcolumntype{L}[1]{>{\raggedright\arraybackslash}p{#1}}
\newcolumntype{C}[1]{>{\centering\arraybackslash}p{#1}}
\newcolumntype{R}[1]{>{\raggedleft\arraybackslash}p{#1}}

\begin{table}[ht]
\caption{Resources that are frequently needed by AI systems.\vspace{-0.3cm}}
\centering
\resizebox{0.9\textwidth}{!}{%
\begin{tabular}{lL{7cm}L{7cm}} \toprule
\textbf{}      &  \textbf{Description}        & \textbf{Example}                                      \\ \hline
\boldmath{$r_d$} & {\em \textbf{Data}}:  All kinds of data (unsupervised, supervised, queries, measurements).  & A self-driving car needs online traffic information.   \\\hline

\boldmath{$r_k$} & {\em \textbf{Knowledge}}:  Rules, constraints, bias, utility functions, etc., that are required.    & A spam filter requires the cost matrix from the user. \\\hline

\boldmath{$r_s$} & {\em \textbf{Software}}:  Main algorithm, associated libraries, operating system, etc.             & A planner uses a SAT solver \inserted{over a complex ecosystem of libraries}.                    \\\hline

\boldmath{$r_h$} & {\em \textbf{Hardware}}:  Computer hardware, sensors, actuators, motors, batteries, etc.           & A drone may need a 3D radar for operation, \inserted{instead of a camera}.          \\ \hline

\boldmath{$r_m$} & {\em \textbf{Manipulation}}:  Manual (human-operated) intervention through assistance                  & A robot needs to be manually re-calibrated \inserted{or overseen real-time}.       \\\hline

\boldmath{$r_c$} & {\em \textbf{Computation}}:  Computational resources (CPU, GPU usage) of all the components           & A \inserted{(vanilla)} nearest neighbor classifier computes all distances.\\\hline

\boldmath{$r_n$} & {\em \textbf{Network}}:  Communication resources (Internet, swarm synchronisation, distribution). & A delivery system \modified{needs online connectivity for} all drones.    \\\hline

\boldmath{$r_t$} & {\em \textbf{Time}}: Calendar (physical) time needed: waiting/night times, iteration cycles. & \modified{A digital assistant} requires cyclical data (weeks) to find patterns. \\ \hline

\boldmath{$r_l$} & {\em \textbf{\insertedTIST{Load}}}: \insertedTIST{Volume, size or dimension of the solution (length of the parameter vector in a DNN, model size in units of bytes, memory usage, etc.)} &  \insertedTIST{A specific DNN (GoogLeNet) trained on CIFAR10 using 8 layers requires 7M parameters and 40MB of storage.} \\ \hline

\boldmath{$r_e$} & {\em \textbf{\insertedTIST{Energy}}}: \insertedTIST{Power consumption per unit of time required to build or operate.} & \insertedTIST{A personal assistant (PA) has a peak power consumption of 2.20W when keyword spotting and 0.4W when idle.} \\

\toprule
\end{tabular}
%\vspace{-0.5cm}
}
\label{tab:resources}
\end{table}
 
%\todo{Table not well motivated... Difference between data and knowledge... (one reviewer says data is a kind of knowledge) One of the reasons they are neglected is that they are conflated... PROPOSAL NOT PERFECT. Data is not data generated by computers, but data from sensors (physical data) or by human behaviour... and before the system operates? So it's not supervision.  Criteria for separation and also for adding to the list....Explain time better.
% more laborious data collection). If so, why is time a separate dimension
% and not part of data? Should then not also the (human or computational) effort
% to gather the data be considered separately?
%Bigger table, difference in cost between different kinds of data or between different kinds of required human manipulation... The lifecycle stage when this is needed. METRICS FOR THE DIMENSION. THIRD COLUMN: possible metrics with references? Difficulty in assigning magnitudes to these dimensions. Ground the ingredients of the dimensions on some deeper understanding of the underlying notions.... economics \url{http://ieeexplore.ieee.org/document/8089878/}.  Say that the framework is general even if some applications will only use a subset of them. One reviewer says $r_m$ is not the same as human-in-the-loop, or supervision...}

\inserted{For some dimensions, we can find methods to evaluate their cost. For instance, software effort can be evaluated using analogy-based, WBS-based or size-based estimation models \cite{putnam1978general,sommerville2007software}. In the hardware category, some models consider both the equipment used for the development and deployment of the system to more complex hardware cost estimation methods, models and tools \cite{ragan2002detailed}. Similarly, compute can be estimated using mathematical (simulation) models such as in \cite{CompCostMath}. In some other cases, these ingredients can be grounded to economic terms \cite{veryard2014economics}, or linked to the concept of ``value proposition'', what a company or product actually delivers to its customers or society % \cite{sales2017ontological}\sidenoteJose{Perhaps a more general source about value propositions?}.} 
\cite{anderson2006customer}.}
%\sidenoteNando{Crees que es apropiado?? Me lo acabo de leer y no me lo parecia. Que tal este: \cite{veryard2014economics}}
%\sidenoteTODO{Read paper and see if it actually has to be included:  \url{http://ieeexplore.ieee.org/document/8089878/}.}} 

\inserted{The ultimate criterion for identifying the resources is that they must incur costs during the development or deployment of an AI system. There are other dimensions that are not necessarily seen as increasing the overall costs, such as {\em fairness}, {\em privacy} and {\em transparency}. Because they have more to do with trust in AI or several ethical issues, they are not included in Table~\ref{tab:resources}. Also,  
they are less neglected nowadays than they were a few years ago \cite{friedler18a,fat2019,fernando2021missing}. %\footnote{\insertedTIST{See \href{http://www.globenewswire.com/news-release/2019/01/17/1701474/0/en/ACM-FAT-CONFERENCE-EXAMINES-FAIRNESS-ACCOUNTABILITY-AND-TRANSPARENCY-OF-ALGORITHMIC-SYSTEMS.html}{http://www.globenewswire.com/news-release/2019/01/17/1701474/0/en/ACM-FAT-CONFERENCE-EXAMINES\\-FAIRNESS-ACCOUNTABILITY-AND-TRANSPARENCY-OF-ALGORITHMIC-SYSTEMS.html}}}\sidenoteJose{Perhaps we can add more references instead of just a link to globenewswire}
In any case, separately or jointly with those in the table, fairness, privacy and transparency could be considered as well when analyzing some particular technologies (especially machine learning), as they can have an impact on % on applicability as   or have to trade off against 
their applicability, if some constraints or regulations are not met, or must be traded off with performance or the other dimensions in Table~\ref{tab:resources}. 
%As prerequisite (privacy) or as a tradeoff with accuracy (fairness and transparency)
In general, as we will see in the following sections, for a particular new innovation or technology, only a subset of dimensions may be relevant.}
It is appealing to collapse \modified{several of these dimensions for} %the benefits and costs of 
an AI system to a single metric. For any given user with rational (transitive and complete) preferences, their preferences can be represented using a utility function. A firm's utility function, for example, might correspond to risk-adjusted expected profit. A user's utility function might be harder to quantify, but is generically increasing in the performance of the system and decreasing in the costs of the system. Denote a performance vector, $\psi$, for a given problem, which is often a unidimensional quantitative score (such as the error), but could also have several components. A utility function maps performance and all associated resources to a single dimension:

\begin{equation}\label{eq:variance}
 \Utility(\psi, \vect{r}) = \Utility(\psi, r_d, r_k, r_s, r_h, r_m, r_c, r_n, r_t, \insertedTIST{r_l, r_e}) \rightarrow \utility
\end{equation}

\noindent In some cases this is an additively separable function, such that  $\Utility(\psi, \vect{r})=\Benefit(\psi) - \sum_x \Cost_x(r_x)$, 
%\sidenoteJose{I introduced notation for costs to make it consistent with the use of costs later on} 
with the first term accounting for the benefit according to the performance of the system minus the costs produced by the use of resources (note that the cost functions $C_x$ are different for each resource). %, such as in 
% (\ref{eq:variance}) can be mapped independently, through particular $\Utility_x(r_x) \rightarrow \utility_x$, whatever resource $x$ is, and performance $\Utility_{\psi}(\psi) \rightarrow \utility_{\psi}$, and added to obtain the total utility. % would be simply $\sum_{x} \utility_x$. 
% %For instance, 
%cost-sensitive classification \cite{elkan2001foundations,turney2002types}.
% JOSE: itis not like cost-sensitive classification: there the costs are not integrated
For economic applications, we might be able to separate the utility function into performance generating revenue (in dollars), and resources imposing costs (in dollars).%\sidenoteAllan{I removed the medical diagnosis problem example as it didn't seem to add much. If add it back in, could be done more concisely.} 
%For instance, in a medical diagnosis problem, we would combine the misclassification costs (in our notation, $\psi$  would represent the false positives and the false negatives, and $\Utility_{\psi}$ %would represent
%the cost matrix) with the attribute costs (%in our notation 
%$r_d$, which would represent the variables, medical tests % that are 
 %needed for the decision, with $\Utility_d$ applying the costs of each test).

\begin{figure}[ht]%
    \centering
    \includegraphics[width=0.4\columnwidth]{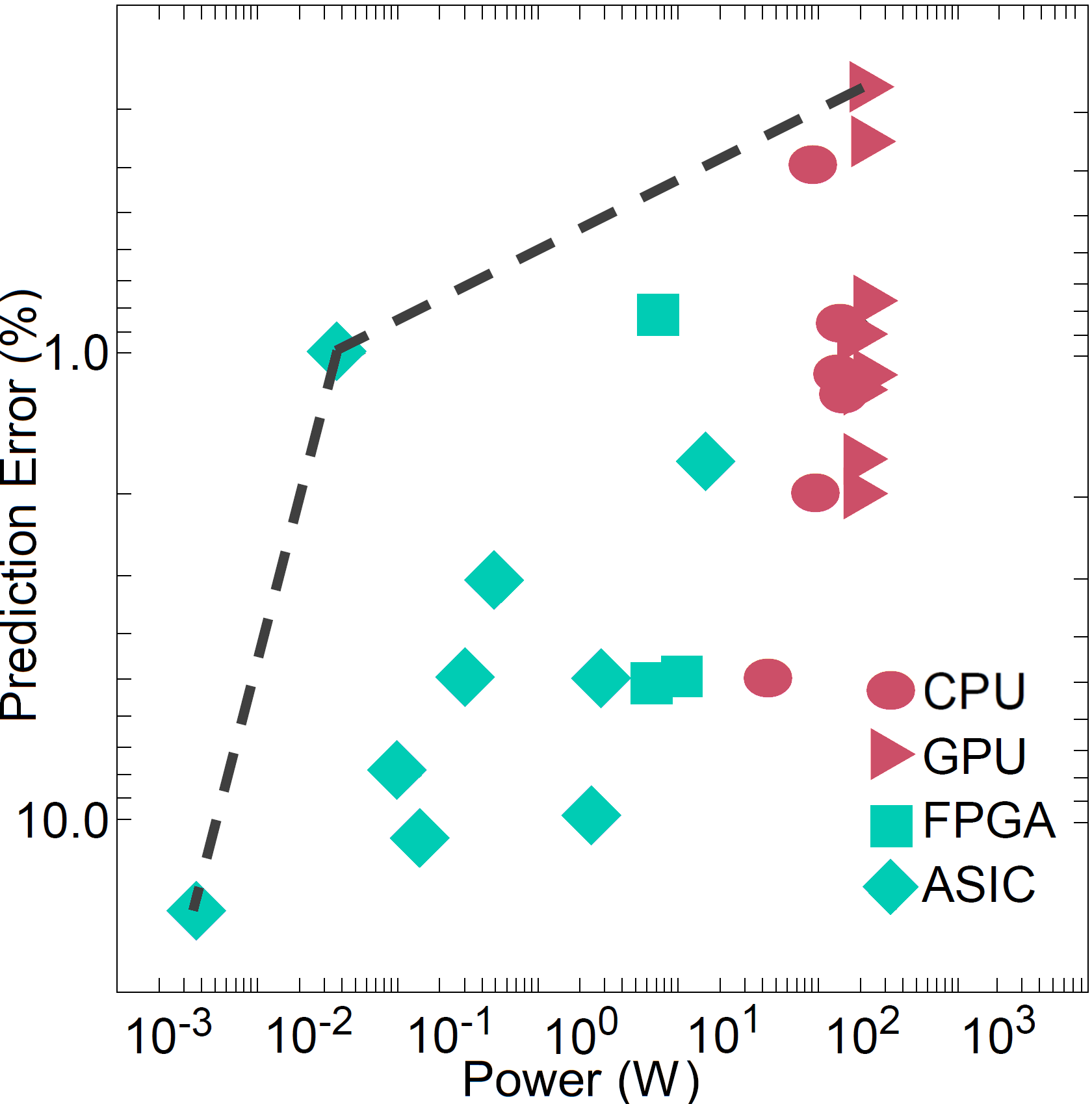}
    \caption{Performance for MNIST \protect\cite{lecun1998gradient}, 
    %a popular handwritten digit recognition benchmark, 
    for 22 papers, 
    compared to power consumption (data from \protect\cite{reagen2017deep}). The Pareto \modified{frontier} is also shown 
    (we will \inserted{later} discuss whether the points can actually be joined by straight segments%in section~\ref{sec:pareto}
    ).}
    %(any two systems can be interpolated by choosing to run one or another with a biased coin). 
%The segments are not straight because of the logarithmic axes.
%}
    \label{fig:pareto1}
\end{figure}

%\todo{Emphasise? One reviewer complains: "So not only do the different dimensions of resources need to be optimized, but the different utility functions of the heterogeneous stakeholders need to be optimized."}
In many cases, we are not able to collapse performance and costs into a single metric, perhaps because the utility function is not known or %the utility function 
varies across a population of users. Still, we can productively examine the relative performance and costs of different systems. For any number of dimensions, we can assess the Pareto-optimal surface. 
%, as we do in Fig.~\ref{fig:pareto1} for two indicators (we explore this further in section~\ref{sec:pareto}\todo{???????????}). We may want to focus on one dimension of costs, such as economic costs or energy costs (as per the ``carbon footprint''). 
For example, 
Fig.~\ref{fig:pareto1} 
%\sidenoteAllan{Nando, if you have time, I think performance should go on y axis, cost on x axis. I also recommend good performance higher up and low energy consumption to the right, so pareto surface is top right. Unless depicting good as bottom left is done in AI papers?}\sidenoteJose{I agree. Nando explained to me, though, that he took the vector image, but not the data, so the only way of doing that would be to draw it from scratch.} 
shows algorithms and architectures according to their MNIST prediction error and power consumption, revealing that most solutions are not on the Pareto surface on these dimensions, with notable exceptions, such as some ASIC architectures, which focus on efficiency in terms of chip space, speed and ``energy footprint'' \cite{chen2014diannao}.
\section{The full range of accounting}\label{sec:ext}

%%%%%%%%%%%%%%%%%%%%%%%%%%%%%%%%%%%%%%%%%%%%%%%%%%%%%%%%%%%%%%%%%%%%%%%%%%%%%%%%%%%%%%%%%%
%%%%%%%%%%%%%%%%%%%%%%%%%%%%%%%%%%%%%%%%%%%%%%%%%%%%%%%%%%%%%%%%%%%%%%%%%%%%%%%%%%%%%%%%%%

%\sidenoteAllan{Last section title was "extent of accountability", but this didn't make sense to me. But this version is not perfect either.}\sidenoteJose{I've further replaced "Towards a full accounting" with "The full range of accounting", as I wanted to show these stages and that we can do the analysis per stages, but we could revert as well.}
The benefits and costs of developing and deploying an AI system are not incurred only once, but throughout the many uses, reuses, and follow-on contributions.  
Some costs %(e.g., compute, data) 
are \modified{borne} exclusively during the initial conception and development, while others recur with each adaptation to a new application, or even each application to a particular user. 
%\sidenoteMiles{maybe "development and deployment" instead of "initial conception and development" as additional compute is used when deployed (not just "developed"?}\sidenoteJose{I've kept it, because the distinction was right, but as you say the example was not, as compute and even data can be used for development and deployment.}
% The previous section has identified the resources that can be used by an AI system. %While they can be analyzed with a utility or cost function, reduced to energy footprint or compared in a Pareto plot, 
% Still, it is important to bound the extent and multiplicities of these resources. For some applications it is important to know whether the computational effort is done during the training or during the deployment of a machine learning model. %For instance, in machine learning, lazy techniques leave all the effort to deployment, while eager techniques anticipate all the efforts. 
% For a particular application, the best choice may depend on the response time needed and how many deployment cases exist, and if the model can be reused. 
In general, the total resource burden should be accounted for according to the whole cycle of the AI system.

\begin{figure*}[!ht]%
    \centering
	\includegraphics[width=1\textwidth]{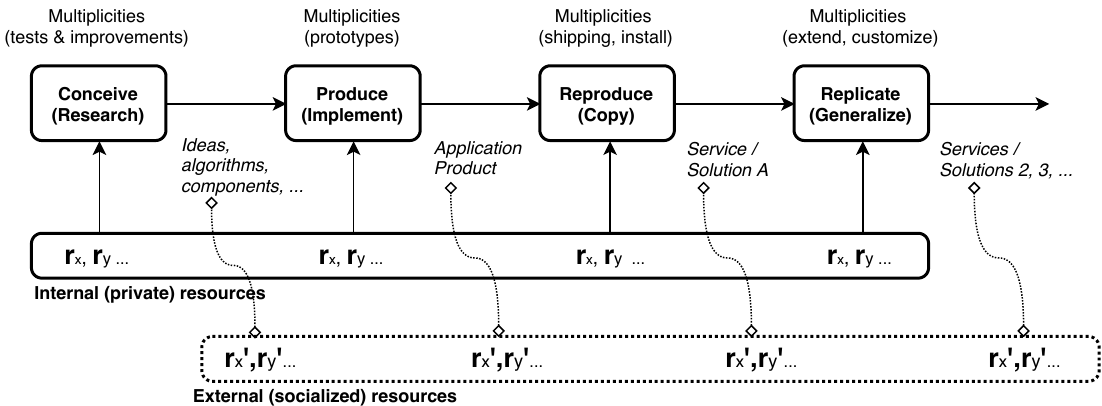}
    \caption{Illustrative representation of  stages of the AI system life cycle where resources might be required.} 
%    This is illustrative, not all AI developments must go through all these stages, or sequentially.}
    \label{fig:IntExtResources}
\end{figure*}

Fig.~\ref{fig:IntExtResources} 
%\sidenoteTODO{DONE-In the figure: Socialised $\rightarrow$ Socialized, and also polish the figure a little bit to fit the space better, and make the text more readable.} 
shows how the dimensions we identified can become relevant at different stages of the life cycle of an AI system. Consider \inserted{we want to assess the potential impact of} a new algorithm for voice recognition. Apart from all the human thinking, there will be a great effort in terms of failed experiments, different libraries used, users testing the early systems, etc. If a company %or start-up 
takes these ideas and builds a prototype, the tests, software, hardware, 
%\sidenote{are we doing Oxford commas or no? If we are, need a comma here}\sidenoteJose{Yes, we're using American English, and Oxford commas are more common there than in England.} 
and compute will concentrate on production. When the system is reproduced (installed or shipped) to users, additional resource costs will be incurred. 
%(with the partial use of computing, hardware, software, etc.). But this is only the start of the story. 
Further, if the idea can be adapted for other applications (e.g., adapting a voice recognition system to  other languages), depending on its generality and reproducibility, the initial contribution can provide further value, at some further adaptation cost including the need for 
% the original can be adapted to these other situations, 
% partially reusing the previous effort,  
% %(perhaps reusing most of the conception, part of the production and most of the reproduction) to these other situations. 
% but needing 
% %Following with the example, adapting the voice recognition system to other languages may require
new corpora, training, semantic knowledge, etc.

%\sidenoteMiles{not sure what proper means here?}\sidenoteJose{Removed}\sidenoteMiles{also, would it be useful (in addition or instead of "socialized") to distinguish between public goods and private goods? or, club goods/other economic concepts? in terms of public goods, some things like frameworks are clearly non-excludible/non-rival, whereas compute is less clear...it's rival because you have to spent money to get it (generally speaking), but potentially open to everyone (to some extent)...and mass use could drive up the cost... though the best computers are off limits to most people (e.g. next gen TPUs).}\sidenoteJose{Not sure what you mean}

%\vspace{-0.7cm}
At each stage of the life cycle, the contribution may be deployed a multiplicity of times  (represented above the boxes in Fig.~\ref{fig:IntExtResources}). The total value of the contribution thus needs to take into account the scale of its deployment. 
% Apart 
% %
% %Clearly, in all these stages there is reuse of many of the resources used in previous stages (e.g., most of the training data may no longer be necessary for the reproduction, and hyperparameter tuning may only be necessary in the first stages). But 
% %Apart 
% from analyzing reusability in this incremental way, we have to pay attention to the multiplicities 
% %\sidenoteJose{Would multiplicities be a better word, also in the figure?}
% involved at each stage (represented above the boxes in Fig.~\ref{fig:IntExtResources}). 
%For every stage, it is not the individual cost of a system that counts but the overall cost of all repetitions, for which there is also reuse. 
%\sidenoteAllan{Jose: what is the system utility, if not the sum of the application utilities? I see in the previous draft we talked about the system cost, vs the application cost. This makes more sense to me. "utility" seems to imply the total benefits and costs of the thing, but the benefit of the "system" includes all the follow-on benefits. I've changed all the following to "costs".}
%\sidenoteJose{Ok, changed notation in the prev ious section too. But now we have to reintroduce the benefit.} 
For instance, some early speech recognition systems were pre-trained once (the \textit{system cost}, denoted by $\systemCost$, 
%\sidenoteJose{Removed subindex for system and application, as we need subindex for the resource, and superindex for application/task. I'm using different fonts. I've defined commands, so this can be changed by other letters: S and A, or sC and aC or some other notation.}, 
covering the ``conceive'' and ``produce'' stages in Fig.~\ref{fig:IntExtResources}) and then adapted to thousands of users, with extra hours of customization per user (the \textit{application cost}, denoted by $\applicationCost^j$ with $j$ indexing each of the $n$ applications, or users, covering the ``reproduce'' and ``replicate'' stages). 
More recent general speech recognition systems do not need such customization. Consequently, the application cost $\applicationCost^j$ is lower per user. 
%\sidenoteAllan{Er, isn't it lower per user?}\sidenoteJose{For utility it was higher, for cost is lower.}
In both cases, the total cost $\Cost$ is $\systemCost + \sum_{j=1}^n \applicationCost^j$. As the number of applications increases, the average cost will converge to the average application cost as the system cost is amortized. For this reason, for contributions that have many possible applications, it is worth paying additional system costs so as to make the contribution more general, adaptable, and reusable, and thereby bring down the application costs.
% Because AI software can be replicated much cheaper than physical products and of course humans, AI research
% %, and AGI research especially, 
% seeks to pay large system costs, in order to achieve low application costs. 
%The possible reuse between stages and multiplicities forces us to be very explicit about those costs that need only be paid once and those that must be repeatedly incurred. 
Since AI often has broad potential applicability, contributions that are general, adaptable, and reusable are likely to have  %especially
high utility, \inserted{so having significant economic and social impact}.%\sidenoteAllan{cut some sentences} 
Fig.~\ref{fig:IntExtResources} not only covers direct ``internal'' costs ($r_x$, $r_y$, \dots) but also some external ``debts'' or %``socialization''
\modified{``societal''} costs ($r_x'$, $r_y'$, \dots). 
%This is sometimes referred to as ``cost externalization''. These costs are usually more difficult to grasp. When they are identified, they sometimes explain some failed AI technology or some global negative impact of the technology. 
For instance, 
%in order to make an AI system work, the developers can reduce resources for production, but they reappear during reproduction or replication. For instance, the early speech recognition systems (Via Voice) have been replaced by more general systems with neglectable application costs (the marginal costs). 
%
%Some other externalities that are more difficult to detect are those that change the service itself (in order to solve the task they change it). A
automated customer service systems (call centers) clearly were not a Pareto improvement relative to previous systems, even though they may be a profit maximizing improvement \inserted{or can represent a baseline of automation for further improvement that is finally assumed as the standard service. In the end,}  % \todo{Reviewer: but can set a baseline of automation for improvement?}. 
%are a good example of this, in which the service looks very different from how it previously operated.
companies reduce their labor costs for customer service by substituting in phone-trees and voice recognition, but in the process impose time, frustration, and other costs onto the customer. 
%\sidenoteMiles{maybe take out the second example so we get the point across without risking offense to people who work on such things? :)}\sidenote{I've added "some", but I've made the point stronger in a way. The goal of these gadgets is to create dependance, as many other commercial products.}
Some navigators and personal assistants can make users more dependent on them, atrophying some %human
capabilities or leading to a simplification of language. 
%\sidenoteMiles{this is fine, made small tweak to grammar}
In other words, the user adapts to the AI system, 
%changing the task, 
and assumes part of the effort or cost. In general, technological innovation both involves developing technology to fit a given conception of the task, and adapting conceptions of the task to fit the capabilities of technology \cite{martinez2021futures,martinez2020tracking}. In the process of adapting work processes, customer expectations, relationship norms, and even urban design to what is technologically convenient, there can be consequences for society that are not internalized by the designers and deployers of these systems. This footprint of AI is not usually acknowledged in benchmarking, \inserted{and can have more societal impact than the technology itself}. 
%the context of AI system evaluation.  
%\sidenoteMiles{though of course some acknowledge it or directly work on such issues - maybe hedge by saying e.g. "not usually acknowledged in the context of AI system evaluations" or something?}
%The impact that maximizing performance at some stages can have at other stages and other actors is more difficult to estimate the wider the window is. While this is the right bird’s-eye view, in the rest of this paper we will focus on more bounded situations, limited to the internal sources only, and how they can be estimated.

From the previous sections, we conclude that the contribution of an AI development should, in principle, be given a full accounting of the costs and benefits, across the contribution's full life cycle. %, % evaluated in terms of all the resources involved, ]
%being explicit about the extent and stages the analysis is performed. 
The current emphasis on targeting and reporting performance benchmarks, however, poses an obstacle  to a full accounting.  
%As argued in section 2\sidenoteJose{argued there, really? Number of the section as text}
Reproducibility and replicability are two traditional tools for addressing this. 
%If we want to evaluate the contribution of a system, idea or algorithm for a single task, then we will reproduce the result and report all the required resources. If we want to evaluate the contribution for a set of tasks (especially if these tasks are variants or related to the original) then we will need to replicate the results for these tasks. 
More precisely: %\sidenoteAllan{These two bullets should be clarified and math updated. I left it because I wasn't exactly sure on what was  meant.}\sidenoteJose{Tried!}

\begin{itemize}%[leftmargin=*,topsep=2pt,itemsep=0pt,parsep=0pt,partopsep=0pt]
\item \textit{Specific reproducibility} refers to whether the \textit{same result} can be obtained from the same conditions and procedures. In AI, this requires that all the necessary code and data are given. This also assumes the same cost functions as well: $\sum_{j=1}^{n} \sum_x \applicationCost_x^j(r_x^j) = n \sum_x \applicationCost_x(r_x)$. 
%\sidenoteAllan{revised slightly, not sure if I distorted meaning. I was unclear what the last line, the cost function statement, meant. Why does this have to be true? Eg, can't we say that your achievement is reproducible, even if it is done by someone with a different cost function?}
%However, if the task is exactly the same, we cannot check generality, beyond the given task. %: $\sum_x \Utility_x(\mu)$. 
\item \textit{General replicability} will check whether the AI technique can be \textit{applied to other problems}, a set of $n$ tasks, applications, %\sidenoteMiles{possible place for Oxford comma (there may be others, too)}\sidenoteJose{ok} 
or users indexed by $j$, with an overall cost $\sum_{j=1}^n \sum_x \applicationCost_x^j(r_x^j)$ that must consider the adaptation effort, with different resources $r_x^j$ and cost functions $\applicationCost_x^j$ per user.  
%positive benefit. As this involves changes, the adaptation effort must be included in 

\end{itemize}

\noindent
%For both reproducibility and replicability, the focus has usually been put on getting “the same results”. But this is not what we mean here. 
Especially for replicability, we can experiment with different hardware architectures, change some of the software and get different computational costs, apart from different 
%key \sidenoteMiles{is the word key needed here?}
performance. That means that the partial results for each $\applicationBenefit^j$ and $\applicationCost^j_x(r_x^j)$ might be different, but we still have something replicable with similar utility. %This makes replicability more open and more difficult (and a lower bound, as some other combinations could have got better results). 
A clear example of this notion of replicability is ``approximate computation'' in deep learning, where one can get much smaller computational costs without a significant change in accuracy 
%, which in some cases falls inside the standard deviation of the variations caused by random weight initialization 
%(see %the ``iso-training noise'' paradigm 
%in %chapter 4 of 
\cite{reagen2017deep}.

%%%%%%%%%%%%%%%%%%%%%%%%%%%%%%%%%%%%%%%%%%%%%%%%%%%%%%%%%%%%%%%%%%%%%%%%%%%%%%%%%%%%%%%%%%
%%%%%%%%%%%%%%%%%%%%%%%%%%%%%%%%%%%%%%%%%%%%%%%%%%%%%%%%%%%%%%%%%%%%%%%%%%%%%%%%%%%%%%%%%%
\section{Exploring the Pareto-frontier of AI research}\label{sec:pareto}
%%%%%%%%%%%%%%%%%%%%%%%%%%%%%%%%%%%%%%%%%%%%%%%%%%%%%%%%%%%%%%%%%%%%%%%%%%%%%%%%%%%%%%%%%%
%%%%%%%%%%%%%%%%%%%%%%%%%%%%%%%%%%%%%%%%%%%%%%%%%%%%%%%%%%%%%%%%%%%%%%%%%%%%%%%%%%%%%%%%%%

%It is easy, if somewhat unpopular, to criticize ongoing practices and demand they be changed. It is much harder to design institutions that can execute such change, while taking into consideration all existing practices, incentives and inertia. In this section we aim to provide initial suggestions about how better surfacing of progress dimensions may be achieved within the AI research community in a sustainable manner.
%\sidenoteJose{I've removed a  paragraph and part of another, as written by Shahar. The first one looked like motivational material for the introduction. This and the first part of the next one went back to a very abstract level. The ideas are right, but it must go more smoothly after the previous section. They are commented and can be recovered.}%
% To start, we must evaluate the intended value of progress measurement. While the sheer thrill of discovery cannot be entirely discounted, we take it to be the primary ultimate aim of AI research to deliver technologies that can benefit society and create value. With this ultimate aim in mind, we need to be mindful of the actors that will be deploying such technologies: corporations, governments, startups, NGOs, and personal users. On the road to achieving benefits, another community of recipients is of special importance: contemporary and future AI researchers. 
Corporations, governments, startups, NGOs, personal users, and contemporary and future AI researchers 
%
%These different actors 
are the intended recipients, or  {\em receivers}, of the AI technologies being developed, % and improved upon, 
and they each have different preferences, resources and constraints, or in other words different operating characteristics \cite{martinez2021research}. The familiar concept of the ROC curve plots true positive rates (TPR) and false positive rates (FPR) for binary classifiers, 
%\sidenoteJose{Now right, but ROC curves are usually TPR against FPR = (1-TNR), and we are using the top left as heaven in many of our plots too.},
and emphasizes the importance of comparing multi-dimensional surfaces, rather than single metrics. %, when comparing technological alternatives. %This has been the main emphasis of our paper throughout, except we have striven to surface a much higher dimensionality of trade-offs faced by receivers of AI technologies. 

\begin{figure}[ht]%
    \centering
	\centering
    \includegraphics[width=0.45\textwidth]{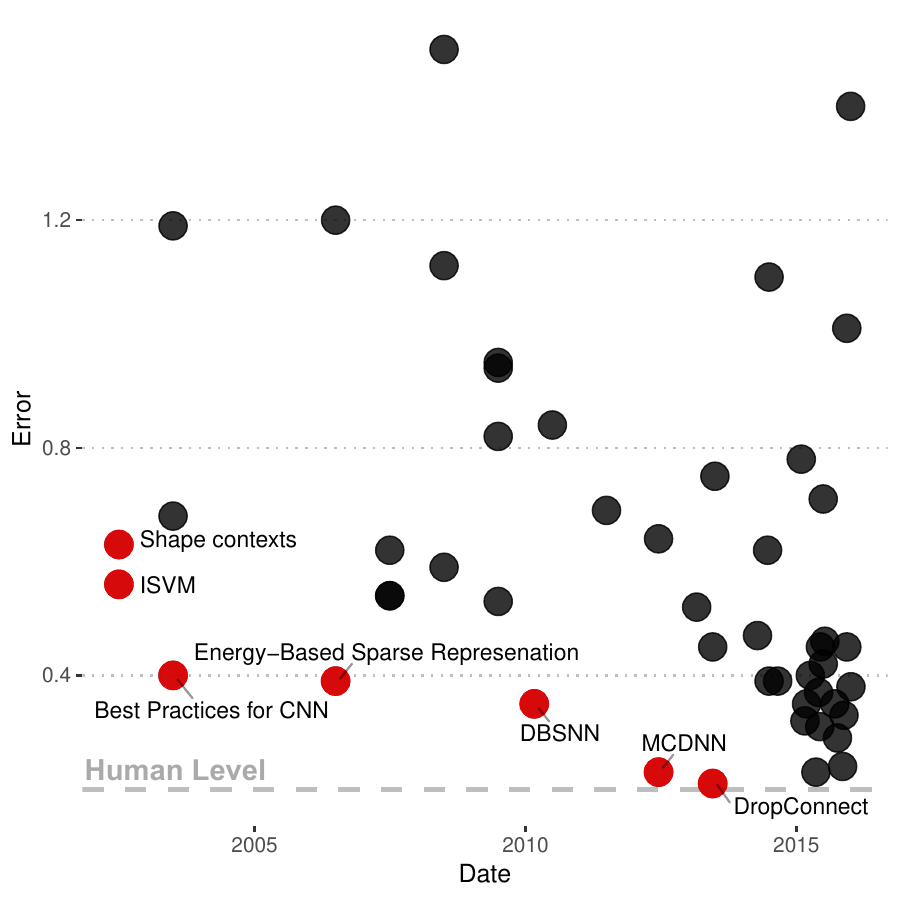}%
    \qquad
    %\enskip
    \includegraphics[width=0.455\textwidth]{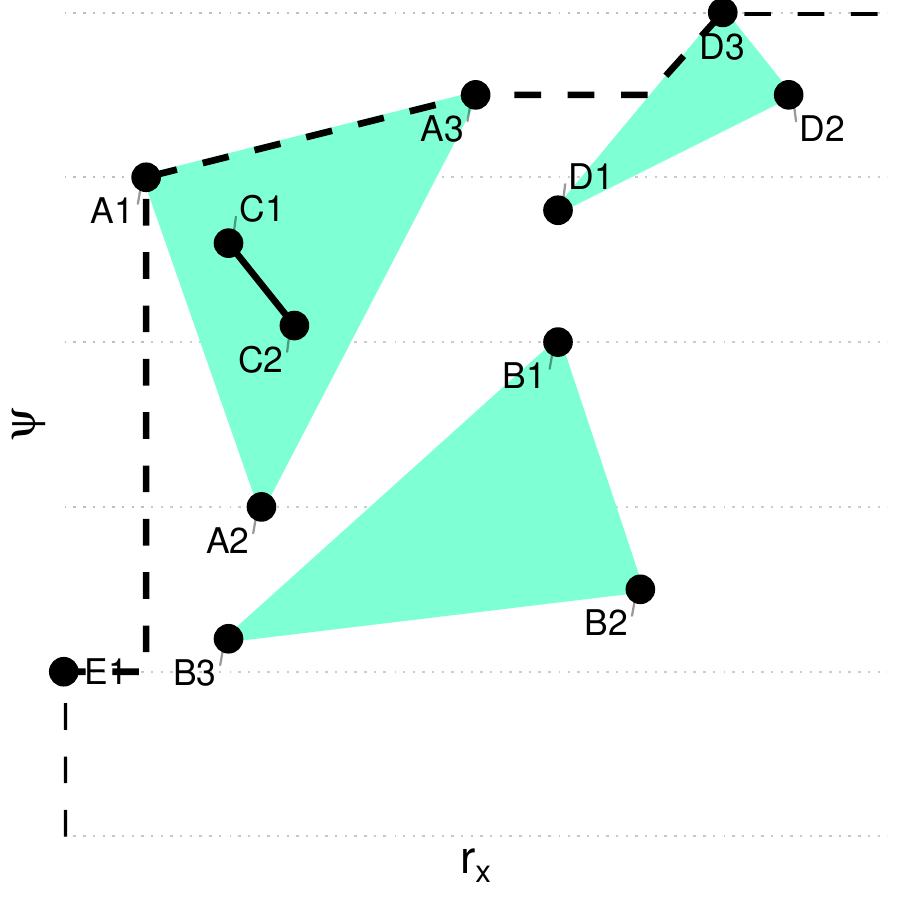}%
    \caption{{Left: Performance for the MNIST benchmark (data from EFF). \inserted{Best state-of-the-art results in red.}} Right: A schematic representation of techniques A, B, C, D, E, with variants, the areas they cover, and the Pareto \modified{frontier} \insertedTIST{(dashsed black line)}.\vspace{-0.3cm}}
    \label{fig:pareto2}
\end{figure}

%\sidenoteJose{This is a (TENTATIVE) last-minute edition, I think it is interesting to make the point of the things that are not visible. Perhaps earlier in the paper, but this is the place where I could fit the figure. }\textcolor{violet}{
For instance, Fig.~\ref{fig:pareto2} (left) just shows a single metric, performance, as a function of time%\sidenoteJose{Are we using red dots when showing state-of-the-art frontier and dashed lines when time is not one of the axes? If so, we should say so}
. This plot does not explain what the cluster of attempts after 2014 really contribute, when they have more error than the already obtained human level. Other dimensions are neglected in this plot, limiting insight about progress. %In the next section we will see other domains where some of the resources are actually put as dimensions. 

Before analyzing the case studies, we have to understand how to build and work with the Pareto \modified{frontier}. When resources are included, the analysis of optimal Pareto surfaces might be slightly different than the traditional triangulization approach. When showing performance metrics such as TPR and FPR for two models, any point in between can be obtained by interpolation, connecting any two points by a straight segment. However, we should note that these points require the implementation of both models. While some of the resources can be interpolated,  
% (as happened with energy in Fig.~\ref{fig:pareto1}), 
others (e.g., software) will simply sum up, and the points between two %other
points will not be achievable with a straight line, but by an axis-parallel route. %This means that, because of some dimensions, we will not be able to apply the convex hull to calculate the Pareto frontier. 
 
For instance,   
Fig.~\ref{fig:pareto2} (right) 
%\sidenoteTODO{done - Make the letters larger so that the figure can be smaller.} 
shows performance against one \modified{hypothetical} resource.  %\sidenoteJose{Shouldn't it be $\Utility(\psi)$ and $\Cost(r_x)$,?}. NOOOPE
For each method, A, B, C, D, and E, the numbers  represent the extremes when varying their parameters. %Similarly for B1, B2 and B3, and D1, D2 and D3. The case with C1 and C2 only allows the modification of one parameter.
E1 represents a random (or baseline) model. Assuming \modified{interpolation is possible by changing the parameters of a method but not possible between different methods,} the Pareto \modified{surface} here is shown in blue. % the areas that can be achieved clearly: moving to the right means more resources and moving down means less performance. 
\modified{Method C can be discarded (as it is covered by A), but method B could also be discarded, as its region is always dominated on the two dimensions by other methods (the dashed Pareto frontier), even if there are unreachable regions in between.} %they do not reach anywhere that cannot already be achieved with A, E and D. %However, we cannot connect E1 and A1, or A3 and D3, or AI and D3 by segments, as they cannot be achieved by interpolation. 

%\todo{One reviewer complains about this figure.}

The diversity of  receivers \moved{---with different subjective utilities---} and the number of dimensions suggest that %the use of %treatment of dimensions as being reducible to 
a single %benefit or 
utility metric is %overly 
simplistic. % ---different receivers would have different subjective utilities %(or, conversely, costs)
%for different dimensions. 
This operating condition  translates into a vector, or gradient, in the multidimensional space. For example, large technology corporations may gain significant utility from a discovery that allows modest speed-ups in exchange for significantly increased compute demands, whereas individual researchers, personal users and startups may find little value in such a discovery. Conversely, the existence of real recipients whose preferences can be known in advance allows us to prioritize exploration of those configurations. 
%the high-dimensional Pareto frontier for AI technologies in only those regions that are occupied, which is expected to be only a small subset of all potential configurations.
%
From the above, we derive a few criteria to identify progress events, \inserted{where all contributions below the Pareto frontier should not be necessarily discarded}: %\sidenoteJose{I would remove this last part of the sentence, from "that merit...".} that merit attention and publication:
%\sidenoteJose{The second and third have been merged into one single bullet point, as Allan and me didn't find them properly distinctive.}

\begin{itemize}%[leftmargin=*,topsep=2pt,itemsep=0pt,parsep=0pt,partopsep=0pt]
\item \textit{Improving the Pareto \modified{frontier} for a known group of recipients} (A1, A3 or D3 in Fig.~\ref{fig:pareto2}, right). This would include all-else-being-equal improvements in performance, but also reductions in computation, data, manipulation %\sidenoteJose{Can we use some of the terms we have in the table here, for consistency?}
or other resources in Table~\ref{tab:resources}. 
This would not, however, 
%include reductions in 
consider extreme
regions 
%dimensions %for which 
no recipient assigns value to. %; thus, improving the data or compute efficiency of approaches that are nowhere near state-of-the-art performance would be of little interest.

\item \textit{Covering a location slightly under the Pareto \modified{frontier} with more flexibility} (B3  in Fig.~\ref{fig:pareto2}, right).
Instead of reaching some areas by combining existing approaches, a new technique can reach there easily with 
a %the use of its parameters. Such advances could show how trade offs can be introduced
trade-off between its own parameters, 
%, e.g., data and compute, or compute and performance,
allowing more receivers to easily find their subjectively optimal trade-offs.% for the domains in which they operate.

\item \textit{Covering a location slightly under the Pareto \modified{frontier} with more diversity} (C in Fig.~\ref{fig:pareto2}, right, if it is very different from A). %, especially if it is near the Pareto-optimal front}. 
The current dominant technique or paradigm can push the Pareto \modified{frontier} for some time, but slightly suboptimal approaches, especially if they are radically different 
%%(Lakatos's 
(i.e., alternative ``research programs''),
%\sidenoteMiles{if we're going to have a philosopher of science reference, maybe say something like "philosopher of science Lakatos's" so people have some context for what we're talking about - don't think most AI researchers would know who this is}\sidenoteShahar{Now that I think of it Lakatos is not the only, or the best, advocate for pluralism in science. I suggest just removing the parentheses. If you really want, you could add a reference, to Lakatos or others (Feyerabend, Laudan, Chang...)}), 
should not be discarded because they may lead to potential improvement in the \modified{frontier} if the current paradigm stalls. %This ``diversity'' progress is Lakatos's alternative ``research programs''. 
%, and also welcoming disruptive vs incremental approaches.
%\sidenoteJose{Sometimes, a small tweak leads to better results with fewer resources and sometimes a completely different approach is only able to reach slightly under state-of-the-art performance. For instance, last year I saw performance on the Atari games using genetic programming  https://web.cs.dal.ca/~mheywood/OpenAccess/open-kelly17a.pdf Multi-task, where deep learning suffers most: https://web.cs.dal.ca/~mheywood/OpenAccess/open-kelly17b.pdf and other things: http://ieeexplore.ieee.org/abstract/document/8085186/) comparable to the Nature paper, but of course too late for being competitive with current state of the art after 1-2 years of progress. Perhaps an element that we should at least mention is diversity, paradigms, etc. Especially for those of you interested in progress, this is a very important consideration (and in the end, diversity has a short-term cost).} 
\end{itemize}

\noindent 
%In order to evaluate the above, it is necessary to hold (at least rough) gradients for different receivers. This, however, is something that r
Receivers can be incentivized to generate and communicate their gradients (though in some cases, countervailing considerations may exist such as commercial secrecy). It is also in the interests of discoverers to %claim and demonstrate the above, either publicly or specifically to
show the recipients benefited by their discovery. Brokers of such information (peer-review, surveys, competitions, etc.) are in a position to meet the incentives (and gradients) of both researchers and recipients by ensuring such discoveries are properly rewarded.%\sidenoteJose{Shahar gave more details.. they are commented out now}
%, when truthful, are published, and are likely to be rewarded (through attention, advertising, higher quality submissions and subscriptions, sponsorship, etc.) for providing such value. It seems that little more is needed than for the brokers of information to take ownership of their responsibility to provide useful, accurate and actionable information to their audiences.\sidenoteShahar{ok, so this is far too Utopian and naive. a full account should consider incentives to cheat, capture of publication venues by powerful actors, under-representation of the needs of weak recipients (NGOs, personal users, heterodox research paradigms), lack of incentives to publish on "external resources". Maybe this should be a footnote expecting and inviting future criticism, while keeping the body relatively upbeat and appealing to journal reviewers?}\sidenote{My take is in between Shahar's and the current text. I think the key norm to change is simply the expectation that scientists should report these dimensions of progress. If scientists expect reviewers to look for them, and others to judge them by whether they included them, then the community will shift. The bad equilibrium arises when no one expects anyone else to care, so no one puts in the effort to report it.}

%%%%%%%%%%%%%%%%%%%%%%%%%%%%%%%%%%%%%%%%%%%%%%%%%%%%%%%%%%%%%%%%%%%%%%%%%%%%%%%%%%%%%%%%%%
%%%%%%%%%%%%%%%%%%%%%%%%%%%%%%%%%%%%%%%%%%%%%%%%%%%%%%%%%%%%%%%%%%%%%%%%%%%%%%%%%%%%%%%%%%
\section{Case studies}\label{sec:case}
%%%%%%%%%%%%%%%%%%%%%%%%%%%%%%%%%%%%%%%%%%%%%%%%%%%%%%%%%%%%%%%%%%%%%%%%%%%%%%%%%%%%%%%%%%
%%%%%%%%%%%%%%%%%%%%%%%%%%%%%%%%%%%%%%%%%%%%%%%%%%%%%%%%%%%%%%%%%%%%%%%%%%%%%%%%%%%%%%%%%%

%\sidenoteTODO{Be more explicit  about compute calculations (see emails). Jose: I guess this has already been done here or in the supplementary material}
%\sidenoteNando{Refer the reader to the supplementary material: \url{https://www.overleaf.com/19060403rdbfbpngfccw#/71395521/}}
In this section we will examine \modifiedTIST{a number of} representative case studies 
%that have recently been referred to as representative examples of significant 
of progress in AI: \modifiedTIST{Alpha*, ALE, imageNet, personal assistants and some others}.

\subsection{Alpha*}
Alpha* refers to a series of papers and associated techniques by DeepMind to play board games. We analyzed the whole series, from  {\sffamily{\small AlphaGo}} \cite{silver2016mastering} (including  the {\sffamily{\small Fan}} and  {\sffamily{\small Lee}} versions, used against Fan Hui and Lee Sedol, respectively, and its latest version, {\sffamily{\small AlphaGo Master}}, which won 60 straight online games against professional Go players), {\sffamily{\small AlphaGo Zero}} \cite{silver2017mastering} (a version created without using data from human games)  and {\sffamily{\small AlphaZero}} \cite{silver2017masteringB} (which uses an approach similar to {\sffamily{\small AlphaGo Zero}} to master not just Go, but also chess and shogi). %These papers combine some classical AI techniques for game playing with deep learning, among other ideas. %We analyse five different methods:
\begin{table}[ht]
\caption{Dimensions (resources and performance) reported in the Alpha* papers. Systems from \cite{silver2016mastering,silver2017mastering,silver2017masteringB}}
\centering
\resizebox{0.9\textwidth}{!}{%
\begin{tabular}{lccccc} \toprule
\textbf{}      &  \boldmath{$AlphaGo_{Fan}$}       &  \boldmath{$AlphaGo_{Lee}$} &  \boldmath{$AlphaGo_{Master}$} &  \boldmath{$AlphaGo_{Zero}$} &  \boldmath{$AlphaZero$} \\ \hline
\boldmath{$r_d$} (Data) & $\checkmark$ & $\checkmark$ & $\checkmark$ & $\checkmark$ & $\checkmark$ \\\hline
\boldmath{$r_k$} (Knowledge) & $\circ$ & $\circ$ & $\circ$ & $\circ$ & $\circ$ \\\hline
\boldmath{$r_s$} (Software) & $\circ$ & $\times$ & $\times$ & $\circ$ & $\times$ \\\hline
\boldmath{$r_h$} (Hardware) & $\times$ & $\times$ & $\times$ & $\times$ & $\times$ \\\hline
\boldmath{$r_m$} (Manipulation) & $\checkmark$ & $\checkmark$ & $\checkmark$ & $\checkmark$ & $\checkmark$ \\\hline
\boldmath{$r_c$} (Computation) & $\checkmark$ & $\circ$ & $\circ$ & $\checkmark$ & $\circ$ \\\hline
\boldmath{$r_n$} (Network) & $-$ & $-$ & $-$ & $-$ & $-$ \\\hline
\boldmath{$r_t$} (Time) & $-$ & $-$ & $-$ & $-$ & $-$ \\\hline
\boldmath{$r_l$} \insertedTIST{(Load)} & $\times$ & $\times$ & $\times$ & $\times$ & $\times$ \\\hline
\boldmath{$r_e$} \insertedTIST{(Energy)} & $\times$ & $\times$ & $\times$ & $\times$ & $\times$ \\\hline
\boldmath{$\psi$} (Performance) & $\checkmark$ & $\checkmark$ & $\checkmark$ & $\checkmark$ & $\circ$
\\\hline

\toprule
\end{tabular}
%\vspace{-0.5cm}
}
\label{tab:Alpha}
\end{table}

%\todo{Explain a little bit why  performance appears in all but one}. 
Table~\ref{tab:Alpha} shows whether the dimensions %that are
were reported in the papers ($\checkmark$), reported in different sources (possibly from different authors) ($\doublecheck$),  only partially accounted for ($\circ$),  not mentioned but relevant ($\times$) and not applicable ($-$).
Many dimensions are relevant for the analysis: the data, the knowledge, the software, the hardware, manipulation, computation and, of course, performance, etc. However, only some of them are provided, which makes a comprehensive comparison of the whole space difficult. Still, we will represent three dimensions: performance (in ELO ranking, which can only be partially estimated for {\sffamily{\small AlphaZero}}), computational resources (using the equivalence: $1\;TPU_{v2} \simeq 3\;TPU_{v1} \simeq 36\;GPU \simeq 180\;CPU$ \cite{jouppi2017datacenter}) %\sidenoteJose{It's somewhat confusing that the xaxis in the figure talks about hardware and we have all crosses in the row table for "hardware". I would clarify the terminology and avoid the use of the word "hardware" when it is not about $r_h$. Talk about CPU GPU speed.}
and human manipulation resources (as represented quantitatively by the ELO ranking of the player or players the system learns from)\footnote{\insertedTIST{Complete information regarding compute can be found in Table \ref{tab:AlphaComp} in the supplementary material.}}. Other dimensions (like knowledge\footnote{{We have the constructed features: stones to be captured or escaped, legal moves, `liberties', etc. While this knowledge is crucial, there is no cost for a new match (reproduction), but the adaptation of {\sffamily{\small AlphaZero}} to other games (replication) may be important}.
%AlphaGo * use some elaborate features such as the liberties, the stones that would be captured or escaped, whether a move is legal, etc.  As for AlphaZero, it needs further knowledge about Chess and Shogi: castling, repetition,no-progress, how pieces move, promotions, piece drops, etc.  While all this knowledge is critical for success, its cost of replication is zero or negligible, but the cost of adapting all this to other games may be very relevant
} 
about Go, % \sidenoteMiles{the software part makes sense to me, but isn't there information about what is/isn't encoded in terms of game knowledge?}\sidenoteNando{``AlphaGo Zero is provided with perfect knowledge of the game rules.''.. "this is the sort of info they provide}\sidenoteMiles{I have a more charitable interpretation of the papers: Extended Data Table 2 in the AlphaGo paper, the "Domain Knowledge" subsection of the Methods section for AlphaGo Zero, and the "Domain Knowledge" subsection of the AlphaZero paper seem to include non-trivial information} 
software, etc.) are not included because of insufficient information from some papers.

%\sidenoteMiles{in the figure, could it be "date published" or "date of publication" instead of "date publication"?}
\begin{figure}[ht]%Alpha_footprint_log
	\centering
    \includegraphics[width=0.8\columnwidth]{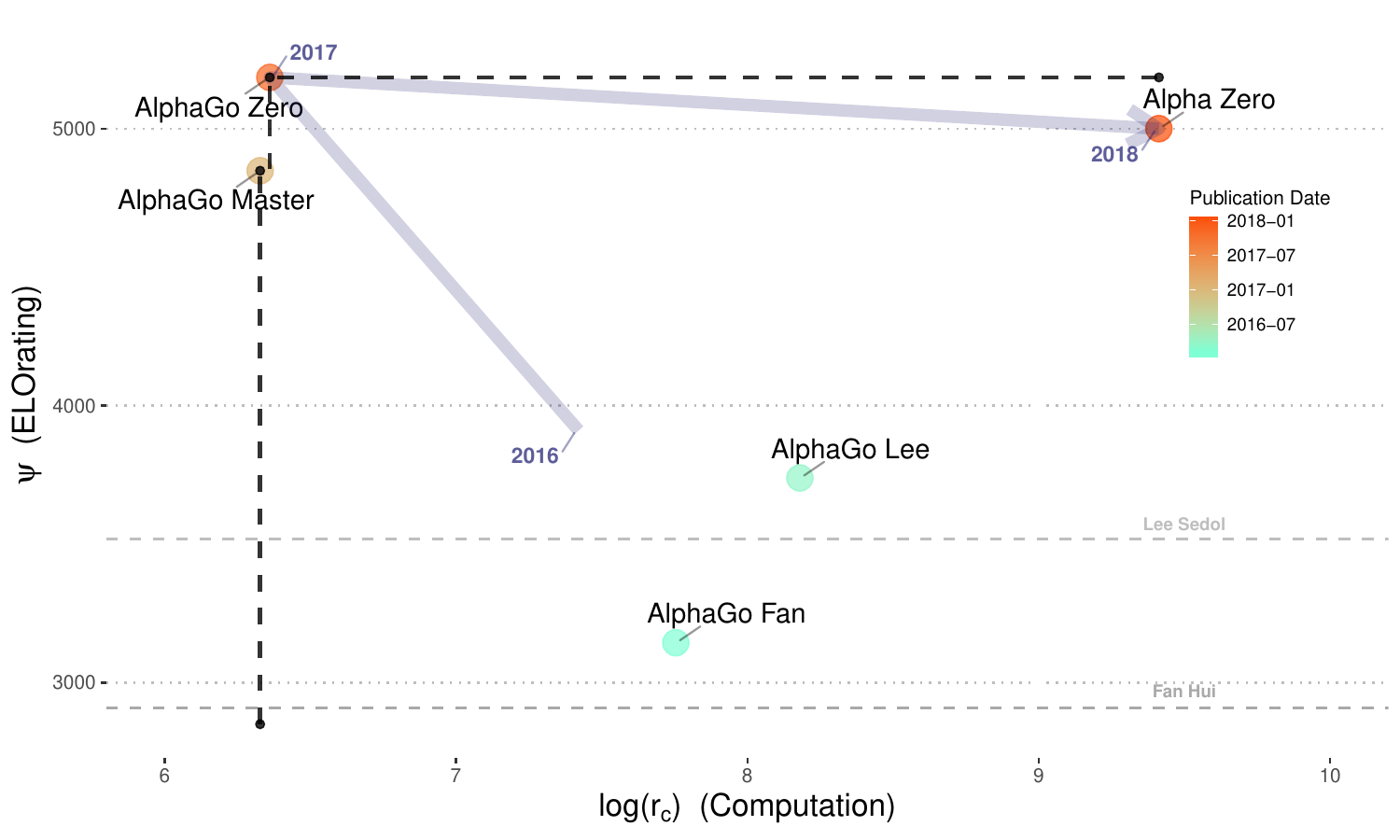}%	
    \caption{Multidimensional utility space for Alpha*. %Pareto-front in green: axis-parallel segments connect systems that cannot be used to generate new ones by interpolation, otherwise, the segments are diagonal. 
    Research gradient evolution from 2015 to 2018 represented with a segmented gray arrow. %Note that while we can apply a convex hull in this case to calculate the Pareto frontier (in marine green), this implies having different systems,
    The Pareto \modified{frontier} (dashed black) does not include other resources % interpolates, so $r_s$ 
    (software, and humans used for training) that duplicate for connecting segments.\vspace{-0.4cm}}%
    \label{fig:AlphaBenchmarks}%
\end{figure}

\begin{figure}[ht]%Alpha_footprint_log
	\centering
    \includegraphics[width=0.8\columnwidth]{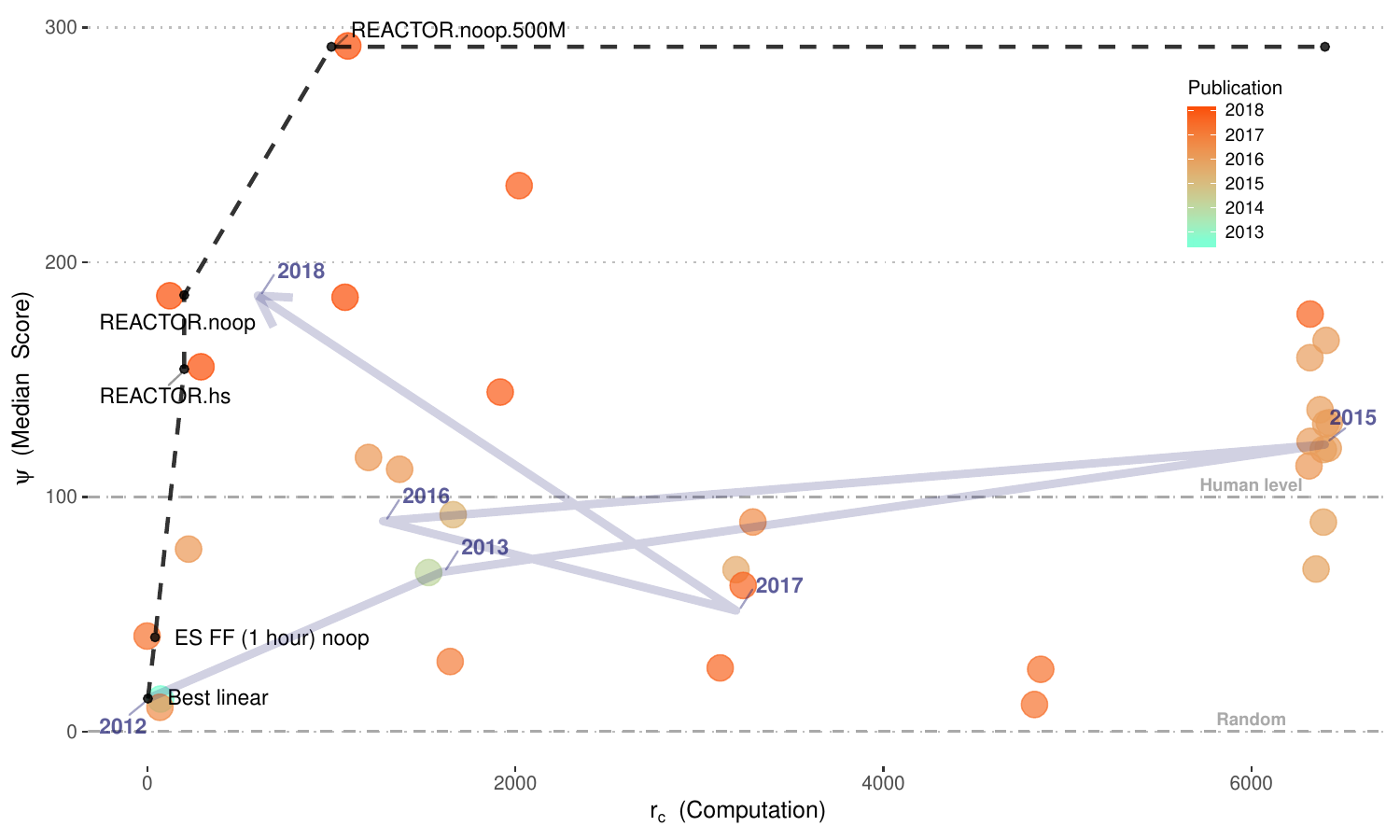}%
    \caption{Multidimensional utility space for ALE. %Pareto-front in green: axis-parallel segments connect systems that cannot be used to generate new ones by interpolation, otherwise, the segments are diagonal. 
    Research gradient evolution from 2013 to 2018 represented with a segmented gray arrow. %Note that while we can apply a convex hull in this case to calculate the Pareto frontier (in marine green), this implies having different systems,
    The Pareto \modified{frontier} (dashed black) does not include other resources % interpolates, so $r_s$ 
    (software, and humans used for training) that duplicate for connecting segments.\vspace{-0.4cm}}%
    \label{fig:ALEBenchmarks}%
\end{figure}

What we see in Fig.~\ref{fig:AlphaBenchmarks} is that the Pareto \modified{frontier} at the moment is represented by {\sffamily{\small AlphaGo Master}} and {\sffamily{\small AlphaGo Zero}}%\footnote{\inserted{Complete information regarding compute can be found in the supplementary material document included with this submission.}}\setcounter{auxFootnote}{\value{footnote}}.
{\sffamily{\small AlphaGo Fan}} and {\sffamily{\small AlphaGo Lee}} are discarded because {\sffamily{\small AlphaGo Zero}} needs less compute\footnote{\insertedTIST{The compute used for generating the training data, i.e. for the self-play games, has not been considered as it is unclear from some of the Alpha* papers (only {\sffamily{\small AlphaZero}}  makes it explicit).}}, \inserted{requires} no %less 
manipulation %(actually none)
and gets better performance. Why is {\sffamily{\small AlphaZero}} \inserted{seen as} a breakthrough if it is not Pareto optimal? The answer is generality.  {\sffamily{\small AlphaGo*}} only solved one task (Go) and {\sffamily{\small AlphaZero}} can solve several tasks. %\sidenoteJose{I don't understand whether something is reused, so we don't get into details for the moment.}\sidenoteNando{According to the authors: ``The AlphaZero algorithm is a more generic version of the AlphaGo Zero''. It differs from the original AlphaGo Zero algorithm in several respects: NN updated continually, invariant to rotation and reflection (training data cannot be augmented), same hyperparameter for all games without game-sespecific tuning, encoded game rules,... \url{https://arxiv.org/pdf/1712.01815.pdf}}  %presumably requires an important adaptation effort (and computational power as it needs  5000 $TPU_{v1}$ to generate self-play games and 64 $TPU_{v2}$ to train the neural networks \cite{silver2017masteringB}) to reuse some of it for other problems \cite{silver2017masteringB}.\sidenoteJose{I thought the only thing it needed was to specify the rules of a new game. That's what Demis said at NIPS.} 
\insertedTIST{Note that the computation times shown in Fig.~\ref{fig:AlphaBenchmarks} include both training and deployment (system and application costs). Hence, a model that is half way between models $A$ and $B$ (choosing between them with equal probability), denoted by $\overline{AB}$, has performance $\psi(\overline{AB}) = 0.5 \psi(A) + 0.5 \psi(B)$, but has a computational cost of $r_c(\overline{AB}) = r_c(A) + r_c(B)$. This is why the Pareto \modified{frontier} in Fig.~\ref{fig:AlphaBenchmarks} has parallel segments, as in Fig.~\ref{fig:pareto2} (right).} Finally, if we look chronologically at the plot, we see that the main gradient that has been followed has been performance.

\begin{table}[ht]
\caption{\modifiedTIST{Dimensions (resources and performance) reported in the ALE papers (from EFF \cite{eff2017} and \protect\cite{Reactor17,Rainbow17})}}
\centering
\resizebox{0.95\textwidth}{!}{%
\begin{tabular}{lcccccccccccccccccccc} \toprule
 &  
 \rotatebox{90}{\boldmath{$Sarsa$}}   &  
 \rotatebox{90}{\boldmath{$Best\; Linear$}}  &  \rotatebox{90}{\boldmath{$DQN\; Best$}} &  \rotatebox{90}{\boldmath{$NatureDQN$}}  &  \rotatebox{90}{\boldmath{$Gorila$}}  &  \rotatebox{90}{\boldmath{$DQN_{noop\;\&\;hs}$}}  &  \rotatebox{90}{\boldmath{$DUEL_{noop\;\&\;hs}$}}  &  \rotatebox{90}{\boldmath{$DDQN_{tuned\;hs}$}} &  \rotatebox{90}{\boldmath{$PRIOR_{noop\;\&\;hs}$}}  &  \rotatebox{90}{\boldmath{$P. DUEL_{noop\;\&\;hs}$}}  &  \rotatebox{90}{\boldmath{$AC3_{LSTM,\;FF,\;FF1d}$}}  &  \rotatebox{90}{\boldmath{$DDQN_{Pop-Art\;noop}$}}  &  \rotatebox{90}{\boldmath{$AC3_{CTS}$}} &  \rotatebox{90}{\boldmath{$Sarsa_{e\;\&\;f-EB}$}}  &  \rotatebox{90}{\boldmath{$TRPO_{hash}$}}  &  \rotatebox{90}{\boldmath{$DQN_{CTS\;\&\;PixelCNN}$}}  &  \rotatebox{90}{\boldmath{$C51_{noop}$}}  &  \rotatebox{90}{\boldmath{$ES\;FF_{(1h)\; noop}$}}  &  \rotatebox{90}{\boldmath{$RAINBOW$}}  &  \rotatebox{90}{\boldmath{$REACTOR$}}\\ \hline
\boldmath{$r_d$} (Data) & $\checkmark$ & $\checkmark$ & $\checkmark$ & $\checkmark$ & $\checkmark$ & $\checkmark$ & $\checkmark$ & $\checkmark$ & $\checkmark$ & $\checkmark$& $\checkmark$ & $\checkmark$ & $\checkmark$ & $\checkmark$ & $\checkmark$& $\checkmark$ & $\checkmark$ & $\checkmark$ & $\checkmark$ & $\checkmark$\\\hline
\boldmath{$r_k$} (Knowledge) & $\circ$ & $\circ$ & $\times$ & $\checkmark$ & $\times$ & $\circ$ & $\times$ & $\circ$ & $\circ$ & $\circ$& $\circ$ & $\circ$ & $\times$ & $\times$ & $\circ$& $\circ$ & $\circ$ & $\times$ & $\checkmark$ & $\checkmark$\\\hline

\boldmath{$r_s$} (Software) & $\times$ & $\times$ & $\times$ & $\checkmark$ & $\times$ & $\times$ & $\times$ & $\times$ & $\times$ & $\times$ & $\times$ & $\times$ & $\times$ & $\times$ & $\times$ & $\times$ & $\times$ & $\times$ & $\checkmark$ & $\times$ \\\hline

\boldmath{$r_h$} (Hardware) & $\times$ & $\times$ & $\times$ & $\times$ & $\times$ & $\times$ & $\times$ & $\times$ & $\times$ & $\times$ & $\times$ & $\times$ & $\times$ & $\times$ & $\times$ & $\times$ & $\times$ & $\times$ & $\times$ & $\times$ \\\hline

\boldmath{$r_m$} (Manipulation) & $\times$ & $\checkmark$ & $\times$ & $\checkmark$ & $\times$ & $\times$ & $\times$ & $\times$ & $\times$ & $\times$ & $\times$ & $\times$ & $\times$ & $\times$ & $\times$ & $\times$ & $\times$ & $\times$ & $\times$ & $\times$ \\\hline

\boldmath{$r_c$} (Computation) & $\circ$ & $\circ$ & $\circ$ & $\circ$ & $\circ$ & $\circ$ & $\circ$ & $\circ$ & $\circ$ & $\circ$ & $\checkmark$ & $\circ$ & $\circ$ & $\circ$ & $\circ$ & $\circ$ & $\circ$ & $\circ$ & $\checkmark$ & $\checkmark$ \\\hline

\boldmath{$r_n$} (Network) & $-$ & $-$ & $-$ & $-$ & $-$ & $-$ & $-$ & $-$ & $-$ & $-$ & $-$ & $-$ & $-$ & $-$ & $-$ & $-$ & $-$ & $-$ & $-$ & $-$ \\\hline

\boldmath{$r_t$} (Time) & $-$ & $-$ & $-$ & $-$ & $-$ & $-$ & $-$ & $-$ & $-$ & $-$ & $-$ & $-$ & $-$ & $-$ & $-$ & $-$ & $-$ & $-$ & $-$ & $-$ \\\hline

\boldmath{$r_l$} \insertedTIST{(Load)} & $\times$ & $\times$ & $\times$ & $\times$ & $\times$ & $\times$ & $\times$ & $\times$ & $\times$ & $\times$ & $\times$ & $\times$ & $\times$ & $\times$ & $\times$ & $\times$ & $\times$ & $\times$ & $\times$ & $\times$  \\\hline

\boldmath{$r_e$} \insertedTIST{(Energy)} & $\times$ & $\times$ & $\times$ & $\times$ & $\times$ & $\times$ & $\times$ & $\times$ & $\times$ & $\times$ & $\times$ & $\times$ & $\times$ & $\times$ & $\times$ & $\times$ & $\times$ & $\times$ & $\times$ & $\times$ \\\hline

\boldmath{$\psi$} (Performance) &$\checkmark$ & $\checkmark$ & $\checkmark$ & $\checkmark$ & $\checkmark$ & $\checkmark$ & $\checkmark$ & $\checkmark$ & $\checkmark$ & $\checkmark$& $\checkmark$ & $\checkmark$ & $\checkmark$ & $\checkmark$ & $\checkmark$& $\checkmark$ & $\checkmark$ & $\checkmark$ & $\checkmark$ & $\checkmark$\\ \hline

\toprule
\end{tabular}
%\vspace{-0.5cm}
}
\label{tab:EFF}
\end{table}

\subsection{ALE}

The second case study %we will analyze 
is ALE \cite{bellemare13arcade}, a collection of Atari games that has become popular for the evaluation of general-purpose RL algorithms learning from screen shots. We selected all the papers (systems) from EFF's AI Progress Measurement Project \cite{eff2017} and the papers introducing \insertedTIST{{\sffamily{\small Rainbow}} \cite{Rainbow17} and {\sffamily{\small REACTOR}}} \cite{Reactor17}. Table~\ref{tab:EFF} shows what information we found about the resources and performance\footnote{\insertedTIST{Complete information regarding compute can be found in Table \ref{tab:EFFComp} in the supplementary material.}}. 
%(same legend as in Table~\ref{tab:Alpha} ), %with a ✓ (the paper includes information about that dimension), o (the paper includes partial information), ✗ (the paper does not include information about that dimension) and − (not applicable).
Again, many dimensions are relevant, but only a few are systematically reported: data, computation and performance. Fig.~\ref{fig:AlphaBenchmarks} represents computation and performance. %\sidenoteJose{Why is data not included?}\sidenoteNando{Where and How? In the plot? How can I include data(games) in the plot?}\sidenoteJose{Ok, we don't have space in any case :-)}
Computation time (whenever the authors do not provide this information explicitly) is roughly estimated from the kind of approach used, whether it is follow-up work, the training setting used, etc., or from figures in more recent papers \modified{making} explicit comparisons between them and the state of the art \cite{Rainbow17,Reactor17}. 

What we see in Fig.~\ref{fig:ALEBenchmarks} is a current Pareto \modified{frontier} dominated by {\sffamily{\small REACTOR}} variants, {\sffamily{\small ES FF}} and {\sffamily{\small Best Linear}}.  \insertedTIST{In this case, the computation times in Fig.~\ref{fig:ALEBenchmarks} includes just training time. If we select a model $\overline{AB}$  that is half way between two models $A$ and $B$ (choosing between them with equal probability),  we may have $A$ train and play for half of the ALE games and $B$ train and play for the rest. As we average for the whole set of games, we can actually have $r_c(\overline{AB}) = 0.5 r_c(A) + 0.5 r_c(B)$, at least if there is no transfer effort between games. This is why the Pareto \modified{frontier} is shown with direct straight segments.}
%If we exclude these latter two because their performance is very poor (but they are very economical in compute), we have that these {\ttfamily{REACTOR}} variants are really recent proposals. 
\modifiedTIST{Regarding the research gradient (in gray), it has changed} over the years, with some disregard of compute initially and more concern in efficiency recently.

For this benchmark, it is common to find ``learning curves'' in the papers (e.g., \cite{machado17}), which %. Learning curves 
show performance varying on %ne parameter, which is 
the number of episodes. This is clearly the $r_d$ (data) but it also influences directly on computation. These learning curves give information of full regions of the multidimensional space, as we saw in Fig.~\ref{fig:IntExtResources}.

For some papers, %have not been included  because %th computation is difficult to compare or 
the comparison was not possible (e.g., due to different subsets of games). It is important to note, however, that some approaches based on genetic programming \cite{kelly} and on planning \cite{bandres2018planning} are valuable in terms of diversity.

\subsection{ImageNet}

\insertedTIST{The third case study is the ImageNet Large Scale Visual Recognition Challenge (ILSVRC) \cite{russakovsky2015imagenet}. Specifically, we focused on the ImageNet classification challenge, a multi-class classification problem framework with about 1.2 million images for training (having 1,000 leaf-node categories in the Imagenet hierarchy), 50,000 for validation and 100,000 images for testing. Each image is associated with one ground truth category, and performance is usually reported as: the top-1 error rate (comparing the ground truth against the first predicted class), and the top-5 error rate (comparing the ground truth against the first 5 predicted classes). Since the breakthrough in 2012 achieved by the first Deep Neural Network (DNN) system {\sffamily{\small AlexNet}} \cite{krizhevsky2012imagenet}, several other DNNs with increasing complexity have been submitted to the challenge in order to achieve better performance. We analyzed the following DNNs which  obtained the highest performance in these past six years, including those more recent approaches developed for environments with a small computational budget (e.g., mobile devices): {\sffamily{\small AlexNet}} \cite{krizhevsky2012imagenet}, {\sffamily{\small BN-AlexNet}} \cite{Zagoruyko2016}, {\sffamily{\small BN-NiN}} \cite{lin2013network}, {\sffamily{\small ENet}} \cite{paszke2016enet}, {\sffamily{\small GoogLeNet}} \cite{szegedy2015going}, {\sffamily{\small VGG}} \cite{simonyan2014very}, {\sffamily{\small ResNet}} \cite{he2016deep}, {\sffamily{\small Inception-v3}} \cite{szegedy2016rethinking}, {\sffamily{\small Inception-v4}} \cite{szegedy2017inception}, {\sffamily{\small Shufflenet}} \cite{zhang1707shufflenet}, {\sffamily{\small Mobilenet-v1}} \cite{howard2017mobilenets}, {\sffamily{\small Mobilenet-v2}} \cite{sandler2018mobilenetv2}, {\sffamily{\small Xception}} \cite{chollet2017xception}, {\sffamily{\small Densenet}} \cite{huang2017densely}, {\sffamily{\small Squeezenet}} \cite{iandola2016squeezenet}, {\sffamily{\small fd-MobileNet}} \cite{qin2018fd}, {\sffamily{\small AmoebaNet}} \cite{real2018regularized}, {\sffamily{\small SENet}} \cite{hu2018squeeze}, {\sffamily{\small Shufflenet v2}} \cite{ma2018shufflenet}, {\sffamily{\small GPipe}} \cite{huang2018gpipe} and {\sffamily{\small PolyNet}} \cite{zhang2017polynet}}.

\begin{figure}[ht]%Alpha_footprint_log
	\centering
    \includegraphics[width=0.9\columnwidth]{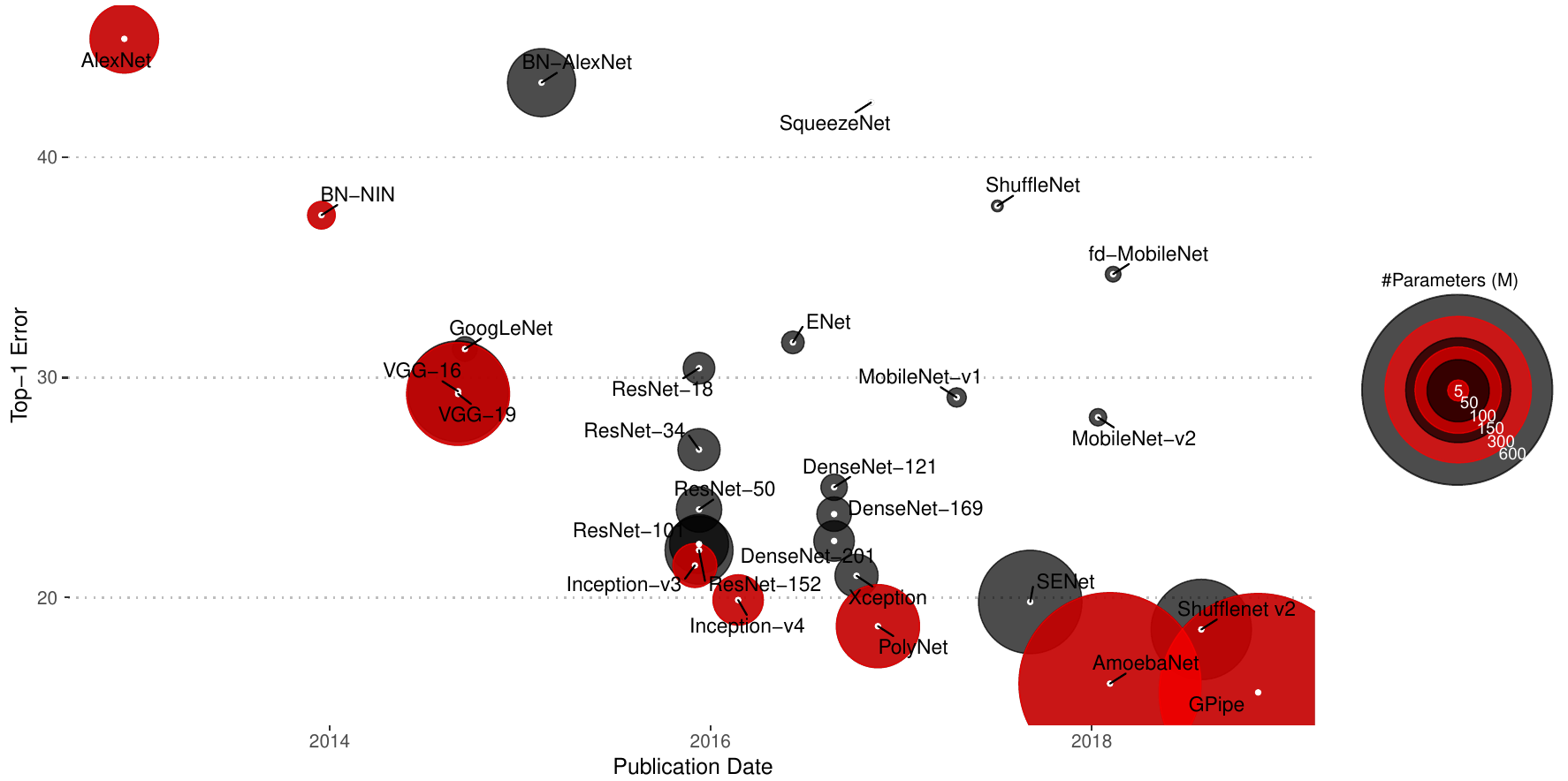}%

    \caption{\insertedTIST{Performance (Top-1 Error) and number of learnable parameters in each DNN for the ImageNet benchmark. Systems are plotted against their publication date.}}%
    \label{fig:imagenetPerformance}%
\end{figure}

\insertedTIST{Table~\ref{tab:ImageNet}}%\sidenoteJose{What does the dou ble tick represent? Say this in the caption. 
\insertedTIST{shows the information we found about the resources and performance\footnote{\insertedTIST{Complete information regarding compute, volume and power consumption can be found in Table \ref{tab:ImageNetComp} in the supplementary material.}}. In this case, many dimensions are relevant and, although some of them are often reported (data, computation, performance, volume), the different hardware/software used as well as the variations in data and training/evaluation techniques precludes a direct comparison of resource utilisation between systems. Furthermore, there is a marked inconsistency in the figures found in  different paper. For instance, for the computation dimension, authors provide either training time, multiply-accumulate (MAC) operations or FLOPs in a single forward pass for input image, etc.,  which are therefore not easily comparable between systems. Still, we can compare some independent dimensions such as \emph{load} (in terms of learnable parameters in each DNN) and performance as a function of time (see Figure \ref{fig:imagenetPerformance}). Again, this plot does not entirely explain what the cluster of attempts after 2017 really contributes to the progress in this domain, when they have more error than the already best result. From the literature we can extract that, in the last few years, authors have increasingly focused on developing highly efficient and accurate networks for very limited computational budgets, which explain why systems such as {\sffamily{\small ENet}}, {\sffamily{\small MovileNet}} or {\sffamily{\small fd-MobileNet}} have reduced their size (compact networks with a significant smaller number of parameters). However, other interesting dimensions are neglected in this plot, limiting insight about progress in this benchmark.}%\sidenoteJose{The big ball used as legend is very confusing, as it looks as another case. Can we plot a square around it or place it outside to make it clear that it is a legend?}

\begin{figure}[ht]%Alpha_footprint_log
	\centering
    \includegraphics[width=0.8\columnwidth]{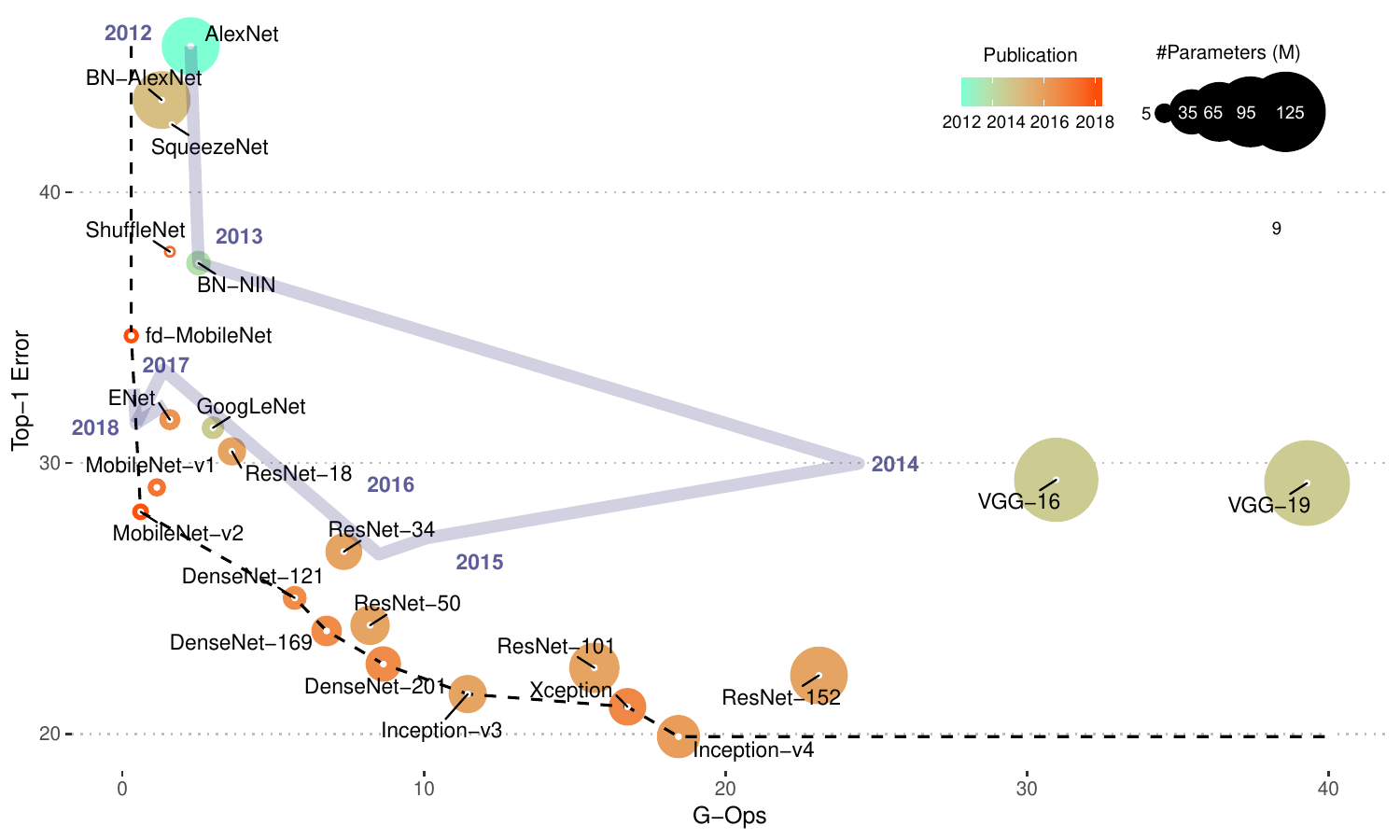}%

    \caption{\inserted{Multidimensional utility spaces for ImageNet. Top-1 Error vs. operations and size of the architectures in terms of millions of parameters. Research gradient evolution from 2012 to 2018 represented with a segmented gray arrow. The Pareto \modified{frontiers} (dashed black) do not include other resources % interpolates, so $r_s$ 
    (software, hardware, etc.) that are duplicated for connecting segments.}}%
    \label{fig:ImageNetPareto}%
\end{figure}

\insertedTIST{In this regard, and given the relevance of the challenge and the wide variety of systems approaching it, some authors have also tried to provide and compare some figures about the quality of different networks in more controlled (and thus comparable) environments in terms of model sizes \cite{real2018regularized}, number of operations (see \cite{real2018regularized,huang2018gpipe}) or evaluation procedures \cite{zhang2017polynet}. A much more comprehensive analysis in terms of computational requirements and performance can be found in \cite{canziani2016analysis}. From the latter, we can obtain metrics related to memory footprint, number of operations and power consumption for a number of systems that can then be used to compare resource utilisation. This can be seen in  Fig.~\ref{fig:ImageNetPareto}, which is a much more insightful and comprehensive plot showing that the current Pareto \modified{frontier} in imagenet is currently dominated by {\sffamily{\small Inception}} variants, {\sffamily{\small Densenet}} and those more efficient networks such as {\sffamily{\small MobileNet}} variants. For the same reason indicated in the previous section for Fig. \ref{fig:ALEBenchmarks}, the Pareto frontier is shown with direct straight segments. The research gradient (in gray) has changed over the years, with some disregard of compute initially and more concern in efficiency recently. Note that not all the papers in Fig. \ref{fig:imagenetPerformance} appear in Fig. \ref{fig:ImageNetPareto}.  For some of them the comparison was neither possible nor acceptable (e.g., mainly due to lack of information but also due to the use of different testing procedures, hardware, software, etc.)}

\begin{table}[ht]
\caption{\modifiedTIST{Dimensions (resources and performance) reported in ImageNet papers (from \cite{krizhevsky2012imagenet,Zagoruyko2016,lin2013network,paszke2016enet,szegedy2015going,simonyan2014very,he2016deep,szegedy2016rethinking,szegedy2017inception,zhang1707shufflenet,howard2017mobilenets,sandler2018mobilenetv2,chollet2017xception,huang2017densely,iandola2016squeezenet,qin2018fd,real2018regularized,hu2018squeeze,ma2018shufflenet,huang2018gpipe,zhang2017polynet}). Note that $\doublecheck$ indicates that a dimension may be reported in sources different from the original (possibly from different authors).}
} 
\centering
\resizebox{0.9\textwidth}{!}{%
\begin{tabular}{lccccccccccccccccccccccccc} \toprule
\textbf{}      &  \rotatebox{90}{\boldmath{$AlexNet$}}   &  \rotatebox{90}{\boldmath{$BN-AlexNet$}}  &  \rotatebox{90}{\boldmath{$SqueezeNet$}} &  \rotatebox{90}{\boldmath{$ShuffleNet$}}  &  \rotatebox{90}{\boldmath{$BN-NIN$}}  &  \rotatebox{90}{\boldmath{$fd-MobileNet$}}  &  \rotatebox{90}{\boldmath{$ENet$}}  &  \rotatebox{90}{\boldmath{$GoogLeNet$}} &  \rotatebox{90}{\boldmath{$ResNet$}}  &  \rotatebox{90}{\boldmath{$VGG$}}  &  \rotatebox{90}{\boldmath{$MobileNet_{v1}$}}  &  \rotatebox{90}{\boldmath{$MobileNet_{v2}$}}  &  \rotatebox{90}{\boldmath{$DenseNet$}} &  \rotatebox{90}{\boldmath{$Inception_{v3}$}}  &  \rotatebox{90}{\boldmath{$Inception_{v4}$}}  &  \rotatebox{90}{\boldmath{$Xception$}} &  \rotatebox{90}{\boldmath{$AmoebaNet$}} &  \rotatebox{90}{\boldmath{$SENet$}} &  \rotatebox{90}{\boldmath{$Shufflenet_{v2}$}} &  \rotatebox{90}{\boldmath{$GPipe$}} &  \rotatebox{90}{\boldmath{$PolyNet$}}\\ \hline

\boldmath{$r_d$} (Data) & $\checkmark$ & $\checkmark$ & $\checkmark$ & $\checkmark$ & $\checkmark$ & $\checkmark$ & $\checkmark$ & $\checkmark$ & $\checkmark$ & $\checkmark$& $\checkmark$ & $\checkmark$ & $\checkmark$ & $\checkmark$ & $\checkmark$& $\checkmark$ & $\checkmark$ & $\checkmark$ & $\checkmark$ & $\checkmark$& $\checkmark$\\\hline

\boldmath{$r_k$} (Knowledge) & $\checkmark$ & $\checkmark$ & $\checkmark$ & $\checkmark$ & $\circ$ & $\checkmark$ & $\checkmark$ & $\checkmark$ & $\checkmark$ & $\checkmark$& $\checkmark$ & $\checkmark$ & $\checkmark$ & $\checkmark$ & $\checkmark$& $\checkmark$ & $\checkmark$ & $\checkmark$ & $\checkmark$ & $\checkmark$& $\checkmark$\\\hline

\boldmath{$r_s$} (Software) & $-$ & $-$ & $\checkmark$ & $\checkmark$ & $-$ & $\checkmark$ & $\checkmark$ & $-$ & $\circ$ & $\checkmark$& $\checkmark$ & $\checkmark$ & $\checkmark$ & $\checkmark$ & $\checkmark$& $\checkmark$ & $-$ & $\checkmark$ & $\checkmark$ & $\checkmark$& $\checkmark$\\\hline

\boldmath{$r_h$} (Hardware) & $\checkmark$ & $-$ & $-$ & $\circ$ & $-$ & $\circ$ & $\checkmark$ & $\circ$ & $\circ$ & $\checkmark$& $-$ & $\circ$ & $\checkmark$ & $\checkmark$ & $\checkmark$& $\checkmark$ & $\checkmark$ & $\checkmark$ & $\checkmark$ & $\checkmark$& $\checkmark$\\\hline

\boldmath{$r_m$} (Manipulation) & $-$ & $-$ &$-$ &$-$ &$-$ &$-$ &$-$ &$-$ &$-$ &$-$ &$-$ &$-$ &$-$ &$-$ &$-$ &$-$ &$-$ &$-$ &$-$ &$-$ &$-$\\\hline

\boldmath{$r_c$} (Computation) & $\circ$ & $-$ & $-$ & $\checkmark$ & $-$ & $\checkmark$ & $\checkmark$ & $\circ$ & $\circ$ & $\checkmark$& $\checkmark$ & $\checkmark$ & $\checkmark$ & $\checkmark$ & $-$& $\checkmark$ & $\circ$ & $\checkmark$ & $\checkmark$ & $\checkmark$& $\circ$\\\hline

\boldmath{$r_n$} (Network) & $-$ & $-$ &$-$ &$-$ &$-$ &$-$ &$-$ &$-$ &$-$ &$-$ &$-$ &$-$ &$-$ &$-$ &$-$ &$-$ &$-$ &$-$ &$-$ &$-$ &$-$\\\hline

\boldmath{$r_t$} (Time)& $-$ & $-$ &$-$ &$-$ &$-$ &$-$ &$-$ &$-$ &$-$ &$-$ &$-$ &$-$ &$-$ &$-$ &$-$ &$-$ &$-$ &$-$ &$-$ &$-$ &$-$\\\hline

\boldmath{$r_l$} \insertedTIST{(Load)} & $\checkmark$ & $\doublecheck$ &$\checkmark$ &$\checkmark$ &$\doublecheck$ &$\doublecheck$ &$\checkmark$ &$\doublecheck$ &$\doublecheck$ &$\checkmark$ &$\checkmark$ &$\checkmark$ &$\checkmark$ &$\checkmark$ &$\doublecheck$ &$\checkmark$ & $-$ & $\checkmark$ & $\checkmark$ & $-$ & $\checkmark$\\\hline

\boldmath{$r_e$} \insertedTIST{(Energy)} & $ \doublecheck$ & $\doublecheck$ &$-$ &$-$ &$\doublecheck$ &$-$ &$\doublecheck$ &$\doublecheck$ &$\doublecheck$ &$\doublecheck$ &$-$ &$-$ &$-$ &$\doublecheck$ &$\doublecheck$ &$-$ &$-$ &$-$ &$-$ &$-$ &$-$\\\hline

\boldmath{$\psi$} (Performance) & $\checkmark$ & $\checkmark$ & $\checkmark$ & $\checkmark$ & $\checkmark$ & $\checkmark$ & $\checkmark$ & $\checkmark$ & $\checkmark$ & $\checkmark$& $\checkmark$ & $\checkmark$ & $\checkmark$ & $\checkmark$ & $\checkmark$& $\checkmark$ & $\checkmark$ & $\checkmark$ & $\checkmark$ & $\checkmark$& $\checkmark$\\\hline

\toprule
\end{tabular}
%\vspace{-0.5cm}
}
\label{tab:ImageNet}
\end{table}

\subsection{Intelligent Personal Assistants}

%%%% Personal Assistants
  
\insertedTIST{Another case study we analyze is AI-powered intelligent personal assistants (PA). We focus on a few big players: {\sffamily{\small Siri}} \cite{Siri}, {\sffamily{\small Alexa}} \cite{Alexa,AlexaSkills}, {\sffamily{\small Cortana}} \cite{cortana} and {\sffamily{\small Google Assistant}} \cite{google}. PAs are mainly based on conversational AI \cite{cassell2000embodied}, natural language processing and knowledge-base systems. They may not represent a `leap' in particular technologies, but an important progression in the integration of the current state-of-the-art techniques to power and improve sophisticated apps and services in terms of latency, automatic speech recognition accuracy, question answering, UI/UX, etc. Within the past few years, all these PAs have been incorporated into a myriad of new physical devices.}

\insertedTIST{However, rarely, if ever, is there evidence regarding how all these PAs have been trained, tested and developed, how knowledge, conversational rules or utterance matching slots are defined, compiled and updated, how data is acquired (from users), which models are trained to understand natural language, reason and interact with human beings, etc. Apart from this, the physical/cloud infrastructure needed is also presumably high with respect to GPU/CPU-based hardware for building the models, data collection/storage/manipulation, testing and scalability tools, architectural and design choices/models, etc.}

\insertedTIST{This is all about internalities, but there are also a number of (neglected) externalities (``societal'' costs) in terms of privacy (e.g., PAs usually collect information about the services that are used (and how and when they are used), or the impact of weariness and distrust towards a sometimes unpolished technology (e.g., misunderstood phrases, incorrect answers and other mistakes, challenging configurations, etc). In this regard, unlike the internalities, there is a number of studies analysing different psychological aspects such as elderly engagement \cite{reis2017using}, user experience \cite{jiang2015automatic}, cognitive workload (mental effort) in voice-based interactions \cite{strayer2017smartphone} or  privacy \cite{Pellungrini:2017:DMA:3167125.3106774,Zhang:2017:EPE:3134224.3078844} and ethical implications \cite{hoy2018alexa,manikonda2018s}.}

\begin{figure}[ht]%Alpha_footprint_log
	\centering
    \includegraphics[width=0.8\columnwidth]{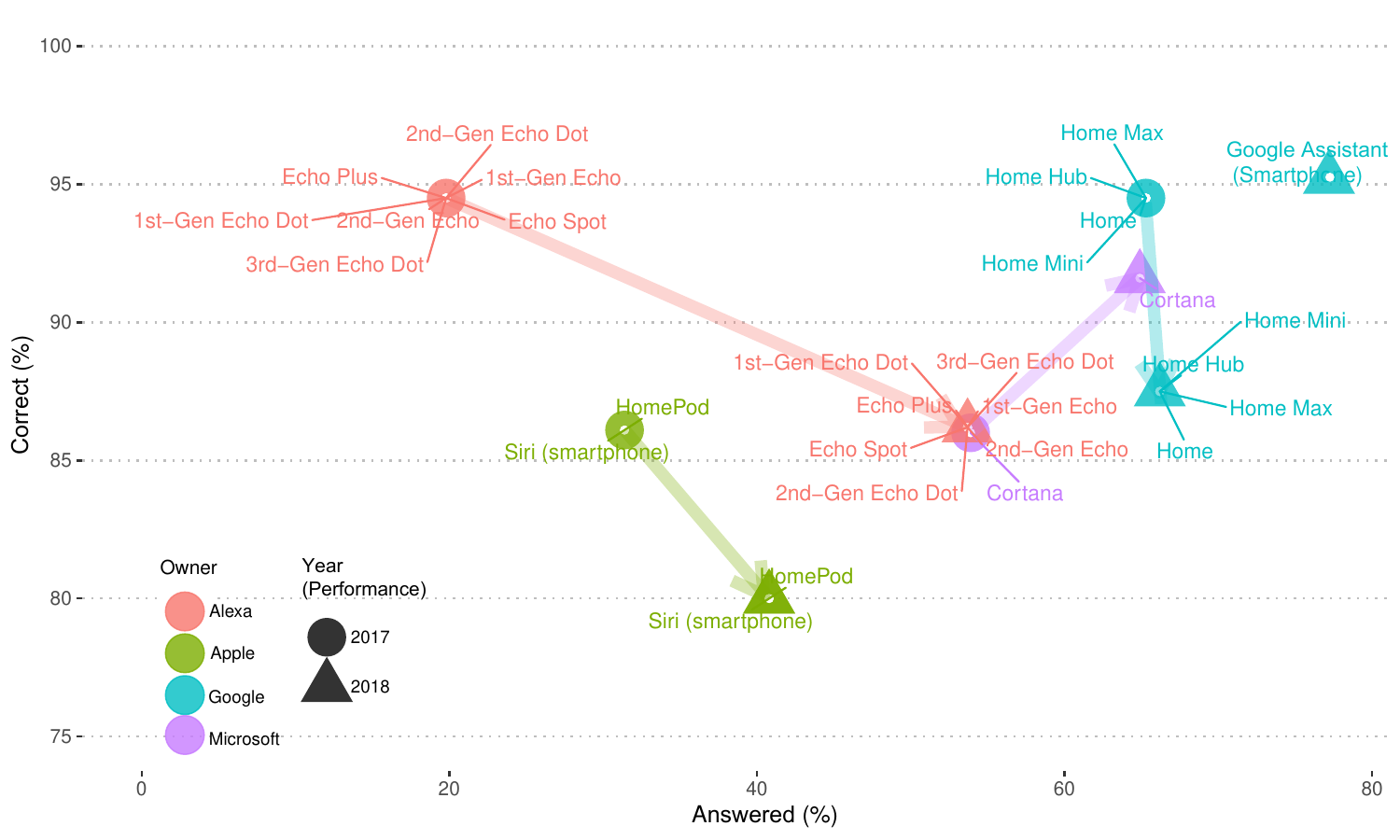}%
 \caption{\insertedTIST{Personal assistants performance over the years 2017 and 2018 \cite{StoneTemple2018}. While \textit{Answered} refer to the percentage of questions in which the PA has attempted to answer obtaining a correct, partially correct or a wrong response, \textit{Correct} refers to the percentage of the questions attempted answered fully correct.}}%
    \label{fig:PAperformance}%
\end{figure}

Focusing on the more neglected internalities, we find that many dimensions are relevant for the analysis: data, knowledge, software, hardware, manipulation, computation, time, load, energy and, of course, performance. However, as we can see in Table \ref{tab:PA}, none of them are directly provided through their documentation (user guides, service manual, datasheets,  websites, etc.) and, when this is the case (e.g., vendor-reported transcription errors \cite{WERgoogle}), these figures are not reliably reported in the literature, or cannot be compared due to companies not following the same standards of evaluation. This means having to make do with external documentation such as reviews, analyses and studies from agencies and other media outlets. 

\begin{figure}[ht]%Alpha_footprint_log
	\centering
    \includegraphics[width=1\columnwidth]{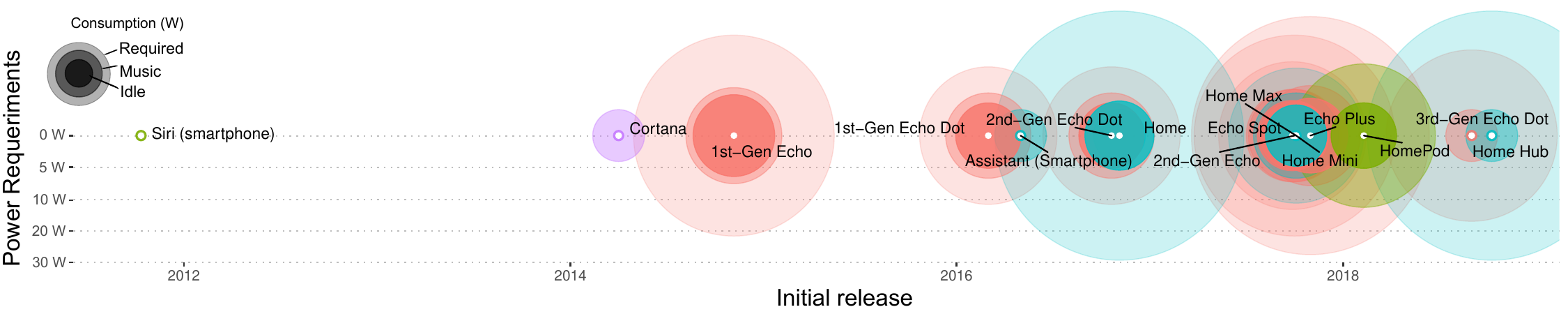}%
 \caption{\inserted{Personal assistant power consumption (minimum power supply requirement, idle and reproducing music).  }}%
    \label{fig:PApower}%
\end{figure}

Accordingly, using several external sources we have been able to obtain information regarding energy requirements and consumption for personal assistants in some scenarios \cite{HTG2018,ambient2018} (Figure ~\ref{fig:PApower}), as well as some performance values from a set of \textasciitilde 5,000 %\sidenoteJose{There is a tilde in the LaTeX code.} 
questions asked to each PA \cite{StoneTemple2018} (Figure ~\ref{fig:PAperformance}). For each question, the authors checked whether the PA answered (i.e., the PA thinks that it understands the question, and makes an overt effort to provide a a correct, a partial correct or an incorrect response to what the user asked for), whether the PA provides a direct and full correct answer to the question answered, %\sidenoteJose{I don't understand the difference between these two, which one is about not understanding the question?}
whether the answer was wrong, and whether the answers were sourced from a database or a third-party source (e.g., Wikipedia). %The experiment was performed twice (in 2017 and 2018) and the results are shown in Figure ~\ref{fig:PAperformance}.

\insertedTIST{What we see in Figures ~\ref{fig:PAperformance} %\sidenoteJose{Are Q Answered correctly or just answered? I think these terms are very confusing... What does it mean to attempt an answer. Can a PA attempt to answer but not answer in the end, even if wrongly?} 
and ~\ref{fig:PApower}} %\sidenoteJose{What is the meaning of "required"??? And why ideal is higher than reproducing music. I don't really understand.The captions should clarify the figures. }
\insertedTIST{ is that, given the incomplete character of the information gathered,  we can only represent the evolution of performance of those different devices/IPAs from 2017 to 2018, as well as the power consumption required for some of them (although the latter is not very illustrative as the energy requirements of these devices are minimal). The most immediate result in terms of performance is that  Microsoft's Cortana outperforms Google on Google Home, where the "Home" version of Google Assistant is not as `smart' as on mobile devices.}%\sidenoteJose{I didn't understand this sentence}
\insertedTIST{Google Assistant for smartphones can be found in 2018 (no data in 2017) the most accurate %smartest\sidenoteJose{I wouldn't use the word "smart" for this} 
PA, attempting to answer %\sidenoteJose{attempting to answer -> answering!} 
almost 80\% of the questions presented with over 90\% in accuracy (correct responses obtained). Cortana also surpasses Alexa and and Apple's Siri by a significant margin.}

\insertedTIST{Regarding the temporal evolution, if we look chronologically at the plot in terms of performance, we see that Amazon Alexa is growing faster than any other PA: while it attempted to answer only 20\% of questions in 2017, in 2018 Alexa attempted to answer over 50\%. 
%main gradient that has been followed has been performance. 
Finally, with respect to energy consumption, although there is an obvious difference of how much electricity various PA devices pull when standby compared to playing music, the average cost per month will not really make a dent in the electricity bill (with costs being under a dollar per month \cite{HTG2018}). However, as they have millions of users, the global impact may be less negligible, especially if we include the consumption on the server side, which is rarely disclosed by the PA companies.}

\insertedTIST{From the these plots (and data), we cannot extract a clear Pareto frontier for any dimensions, or the research gradient over the years, making it difficult (if not impossible) to assess the different contributions or their economic and social impact. Having into account that these systems are thought to be ubiquitious in the future, this at least worrying, and more effort (or regulations) should be done so that some of the dimensions could be analysed and compared by users, governments and regulators.}%\sidenoteJose{The analysis here is perhaps too negative and convoluted. Ok, but perhaps it can be streamlined a little bit.}

\begin{table}[ht]
\caption{\modifiedTIST{Dimensions (resources and performance) for virtual personal assistants: {\sffamily{\small Siri}} \cite{Siri}, {\sffamily{\small Alexa}} \cite{Alexa,AlexaSkills}, {\sffamily{\small Cortana}} \cite{cortana} and {\sffamily{\small Google Assistant}} \cite{google}.}
} 
\centering
\resizebox{0.6\textwidth}{!}{%
\begin{tabular}{lccccccccccccccccccccccccc} \toprule
\textbf{}      &  \rotatebox{90}{\boldmath{$Amazon Echo Dot$}}   &  \rotatebox{90}{\boldmath{$Amazon\; Echo$}}  &  \rotatebox{90}{\boldmath{$Amazon\; Echo\; Plus$}} &  \rotatebox{90}{\boldmath{$Amazon\; Echo\; Spot$}}  &  \rotatebox{90}{\boldmath{$Google\; Home$}}  &  \rotatebox{90}{\boldmath{$Google\; Home\; Mini$}}  &  \rotatebox{90}{\boldmath{$Google\; Home\; Max$}}  &  \rotatebox{90}{\boldmath{$Google\; Home\; Hub$}} &  \rotatebox{90}{\boldmath{$Appe\; HomePod$}} &  \rotatebox{90}{\boldmath{$Apple\; Siri$}}  &  \rotatebox{90}{\boldmath{$Google\; Assistant$}}  &  \rotatebox{90}{\boldmath{$Microsoft\; Cortana$}}\\ \hline

\boldmath{$r_d$} (Data) & $\times$ & $\times$ & $\times$ & $\times$ & $\times$ & $\times$ & $\times$ & $\times$ & $\times$ & $\times$& $\times$ & $\times$\\\hline

\boldmath{$r_k$} (Knowledge) & $\times$ & $\times$ & $\times$ & $\times$ & $\times$ & $\times$ & $\times$ & $\times$ & $\times$ & $\times$& $\times$ & $\times$\\\hline

\boldmath{$r_s$} (Software) & $\times$ & $\times$ & $\times$ & $\times$ & $\times$ & $\times$ & $\times$ & $\times$ & $\times$ & $\times$& $\times$ & $\times$\\\hline

\boldmath{$r_h$} (Hardware) & $\times$ & $\times$ & $\times$ & $\times$ & $\times$ & $\times$ & $\times$ & $\times$ & $\times$ & $\times$ & $\times$& $\times$\\\hline

\boldmath{$r_m$} (Manipulation) & $\times$ & $\times$ & $\times$ & $\times$ & $\times$ & $\times$ & $\times$ & $\times$ & $\times$ & $\times$& $\times$ & $\times$\\\hline

\boldmath{$r_c$} (Computation) & $\times$ & $\times$ & $\times$ & $\times$ & $\times$ & $\times$ & $\times$ & $\times$ & $\times$ & $\times$& $\times$ & $\times$\\\hline

\boldmath{$r_n$} (Network) & $-$ & $-$ &$-$ &$-$ &$-$ &$-$ &$-$ &$-$ &$-$ &$-$ &$-$ &$-$\\\hline

\boldmath{$r_t$} (Time) & $\times$ & $\times$ & $\times$ & $\times$ & $\times$ & $\times$ & $\times$ & $\times$ & $\times$ & $\times$& $\times$ & $\times$\\\hline

\boldmath{$r_l$} \insertedTIST{(Load)} & $\times$ & $\times$ & $\times$ & $\times$ & $\times$ & $\times$ & $\times$ & $\times$ & $\times$ & $\times$& $\times$ & $\times$\\\hline

\boldmath{$r_e$} \insertedTIST{(Energy)} & $\circ$ & $\circ$ &$\circ$ &$\circ$ &$\circ$ &$\circ$ &$\circ$ &$\circ$ &$\checkmark$ &$\times$ &$\times$ &$\times$\\\hline

\boldmath{$\psi$} (Performance) & $\doublecheck$ & $\doublecheck$ & $\doublecheck$ & $\doublecheck$ & $\doublecheck$ & $\doublecheck$ & $\doublecheck$ & $\doublecheck$ & $\doublecheck$ & $\doublecheck$& $\doublecheck$ & $\doublecheck$ \\\hline

\toprule
\end{tabular}
%\vspace{-0.5cm}
}
\label{tab:PA}
\end{table}

%\todo{Comment something on other benchmarks? IBM Watson?}

\subsection{Others}

%\sidenoteJose{This makes sense if we can say something very brief about some other domains, so that "Others" really make sense. For instance, we could say something about self-driving cars and the cost of sensors: radar vs. cameras..., and how obscure the domain is, given the complexity of the product, the regulations and the relevance of marketing and industrial secrets and strategies of the car manufacturers.}
\insertedTIST{Further case studies can be analyzed, although with even much less detail and absence of data, so we just briefly comment on some important cases that are considered influential or impactful in AI.}

\insertedTIST{IBM Watson \cite{Ferrucci:2012}, which uses cognitive computing technology together with information retrieval support and natural language processing, has been successfully applied in different domains, such as health care and life sciences, education or business analytics. However, it presumably needs a large degree of adaptation effort across domains. It is not always explicit how much this effort is, as this is also part of the business model for IBM. As in many other commercial products, full replicability is not possible for obvious reasons, and only some general architectural aspects of IBM Watson are known.}

\insertedTIST{Another interesting (yet again obscure) domain to explore is self-driving cars. Here,  reliability, safety \cite{hernandez2019surveying,hernandez2020ai} %appearance\sidenoteJose{I don't know what you mean with appearance}\sidenoteNando{\url{https://dl.acm.org/citation.cfm?id=3131726.3131750}} 
and social acceptance are only a few of the most well-known costs, but there are a lot more. Autonomous vehicles are not single devices but a collection of hardware and software pieces \cite{wei2013towards} applied in complex and novel ways, so involving new research, development and production costs. For instance, with regard to hardware, % involves from sensors to radars and cameras, with radically different costs: 
while radars and sensor are already cheap and robust enough to be incorporated into mass-market cars, laser-shooting \textit{LIDAR} or 3D photometry are still expensive solutions. 
%, requiring substantial R \& D and testing yet.
The same happens with software: safe driving requires more than state-of-the-art computer vision, it is also necessary to identify blind spots, use artificial intuition, anticipate driving behaviour, etc. All of this requires breakthrough end-to-end solutions with high costs in terms of data collection, model training, and testing in both  simulation and real environments. Again, it is not always clear and transparent how much of this effort comes from high-level competition in the market between these players %the  major players\sidenoteJose{Do we need to name the major players? Are we sure these are the major players?}
(e.g, Waymo\footnote{\url{https://waymo.com/}}, General Motors\footnote{\url{https://getcruise.com/}}, Uber\footnote{\url{https://www.uber.com/info/atg/technology/}}, Tesla\footnote{\url{https://www.tesla.com/autopilot}}, etc.). On the other hand, we can also think about the not fully-explored societal costs (or benefits) self-driving cars might entail in terms of unemployment\cite {poczter2014google}, pollution \cite{selfdrivingPollution}, traffic \cite{selfdrivingTraffic}, infrastructures and urban design \cite{urmson2008self}, fatalities \cite{sivak2015road}, etc. In the same way, regulations are now compulsory for cars in terms of declaring their emissions, some other indicators of the AI self-driving technology that is incorporated in a car should also be disclosed and approved, so that we could plot several dimensions. Otherwise the cars of the future may end up having very efficient engines, but very inefficient `brains'.}

%  \sidenote{
% Other sources for examples or benchmarks:
% o   https://www.eff.org/ai/metrics
% o   https://srconstantin.wordpress.com/2017/01/28/performance-trends-in-ai/   (most included above)
% o   https://aiindex.org/2017-report.pdf ("Technical Performance" and "Towards Human-Level Performance?" are the most relevant parts.)
% o   imagenet
% http://dawn.cs.stanford.edu/2017/11/29/dawnbench-intro/ 
% Baidu paper:  https://arxiv.org/abs/1712.00409 
% Another possible thing to mention here: https://arxiv.org/abs/1801.06146.
% }

%%%%%%%%%%%%%%%%%%%%%%%%%%%%%%%%%%%%%%%%%%%%%%%%%%%%%%%%%%%%%%%%%%%%%%%%%%%%%%%%%%%%%%%%%%
%%%%%%%%%%%%%%%%%%%%%%%%%%%%%%%%%%%%%%%%%%%%%%%%%%%%%%%%%%%%%%%%%%%%%%%%%%%%%%%%%%%%%%%%%%
\section{Conclusions}\label{sec:conc}
%%%%%%%%%%%%%%%%%%%%%%%%%%%%%%%%%%%%%%%%%%%%%%%%%%%%%%%%%%%%%%%%%%%%%%%%%%%%%%%%%%%%%%%%%%
%%%%%%%%%%%%%%%%%%%%%%%%%%%%%%%%%%%%%%%%%%%%%%%%%%%%%%%%%%%%%%%%%%%%%%%%%%%%%%%%%%%%%%%%%%

%\todo{Rewrite saying that concerns were shared in other areas before, and cost-sensitive analysis is an example...COMMENT: the data necessary for the proposed framework was not readily available in the publications that they considered as case studies. Applying the proposed framework is impossible for current and past approaches because of a lack of data, and would only be made possible if researchers put all the required information in future publications or a central database. It is very questionable whether this would work in practice, given the amount and granularity of data required.
% ANOTHER COMMENT: "benchmarking research can be counterproductive, and create resistance"... We're measuring products that need to be deployed and applied to the application but to many other societal needs. Only in this way can we really measure their impact. Also many senior researchers in AI are aware of many of these concerns... But his produces a reference / systematic way that they can use to suggest other authors and reviwers to ahve these things considered more prominently...}

The interest in more comprehensive evaluation protocols \inserted{to assess the potential impact of new AI technologies and the progress they represent}, going beyond performance alone, \modified{was illustrated} by some of the references we included in \modified{the background section}. %\ref{sec:back}}
%\todo{MISSING LINK! Rewrite?} 
%on cost-sensitive learning, reproducibility, generality, data-efficiency and computational costs. 
%
%However, only by an explicit enumeration of all the dimensions (as represented by Table~\ref{tab:resources}) and their integration into utility functions or representation in a multidimensional space, with a clear delimitation of the extent of accounting, can we formalize this analysis, so that a novel contribution can be rigorously evaluated as a contribution to AI progress more broadly. 
%
However, in order to rigorously evaluate \modified{the impact a new contribution in AI can have more broadly, 
%a novel contribution to AI progress more broadly,
%we need a more formal analysis. This is done by 
we need} an explicit enumeration of all the dimensions (as represented by Table~\ref{tab:resources}) and their integration into utility functions or their representation in a multidimensional space, with a clear delimitation of the extent of accounting. \inserted{This represents a novel model to help anticipate the impact of a particular AI technology, bringing dimensions that are usually part of other disciplines or still not sufficiently technical or developed to be considered as parameters to optimize or evaluate. The several scenarios we have analyzed in this paper illustrate how the evaluation techniques can be applied in practice, but they also show that more transparency (through accountability, openness and replicability) has to be applied to AI research in the first place, to assess the contributions and potential economic and social impact more scientifically.}
%\sidenoteJose{This new text tries to capture some of things that will be evaluated by the reviewers, according to the criteria here: \url{https://aaai.org/Conferences/AAAI-19/aaai19emergingcall/}}

\inserted{Of course, there can be resistance from AI researchers and reviewers, as more dimensions and indicators in papers, products and competitions can be seen as a counter-productive burden. Also, the lack of these dimensions in many papers today make illustrative examples, such as those we have included here, more challenging.}

\inserted{While we share some of these concerns, we have to look retrospectively to areas that were completely neglected a few years ago \cite{hager2017advances}, such as fairness, with a wide range of technical metrics that can be used in utility functions or in trade-offs against other dimensions. Similarly,}  this is what happened in cost-sensitive learning more than 15 years ago \cite{elkan2001foundations,turney2002types}, leading to a wide range of techniques that covered different operating conditions. % (e.g., decision trees only using three medical tests for a diagnosis were preferable for a medical context than more powerful techniques using all attributes).
While all these costs are nowadays integrated into the measures of performance, many other resources are not, as we have surfaced here. 
%We hope this paper can launch the study of  “cost-sensitive AI”. 
%
Within this framework, %we have presented here, 
we make a series of recommendations:

\begin{itemize}%[leftmargin=*,topsep=2pt,itemsep=0pt,parsep=0pt,partopsep=0pt]
\item Benchmarks and competitions should be defined in terms of a more comprehensive utility function, considering as many dimensions as possible, or %still better, 
recognize the value of all contributions that have any of the positive effects on the Pareto \modified{frontier} identified \modified{previously}%in Section \ref{sec:pareto}
, in short or long terms. 

\item Papers presenting or evaluating algorithms should generally %\sidenoteMiles{weakened slightly again}
try to report the whole region they cover, % (or some contouring points from which to interpolate), 
and how to navigate the region by modifying parameters or resources. There are many partial examples nowadays: learning curves, plots comparing the number of models vs. performance, planning performance vs. lookahead, etc.

\item These utility functions and multidimensional spaces must also be seen in terms of replicability, for variants of the problems and at different stages of the AI life cycle. %Here the graphical representation is more difficult, as 
The multiplicities %(repetitions) are included,
are more difficult to plot graphically, but we can still define operating conditions depending on the adaptation (or transfer) effort for $m$ problems, or $n$ users.

\end{itemize}

\noindent Frequently, we %will see that
%paper A is not better than paper B. It is just that they
 will not be able to say that one technique is `better' than another: they just cover different regions of the multidimensional space. It is the receiver who will choose the system that best fits their needs. Having a representation of the Pareto \modified{frontier} may hugely facilitate this choice for other researchers and industry, as simply as moving the gradient  until touching the Pareto surface. Also, small players in AI could focus on those areas that require less resources and still contribute to the Pareto \modified{frontier} or to diversity. Finally, the Pareto surface can help detect some societal risks, \inserted{and unexpected huge social impact},  especially if we see that a powerful capability in AI can be achieved with very few resources, becoming available to malicious actors.  

This view of the operating condition as a gradient may suggest clever approaches to push the \modified{frontier} for some resources, as gradient descent is increasingly being used at a meta-level \cite{Andrychowicz16}. In general, we hope this paper will help change perceptions, promote more general and versatile techniques, highlight the trade-offs, and raise awareness of the overall ``AI footprint'', well beyond performance. %not only beyond performance, but also beyond the internal costs.
%\sidenoteNando{\cite{aghion2017artificial,korinek2017artificial} cited in "Background"Components and Integration" ("automation" paragraph)}
%\sidenoteNando{\cite{hager2017advances} cited in "Conclusions"}
%\sidenoteNando{\cite{hwang2018computational} cited in "Background"}
%\sidenoteNando{"pont2013measures" WAS cited in the "Intro" (meta-measures), now removed}
%\sidenoteNando{"sales2017ontological" not used. Substituted by a  more general cite \cite{anderson2006customer}}

%\todo{Read and potentially comment on some of these papers:
%\cite{aghion2017artificial,hager2017advances,hwang2018computational,korinek2017artificial,minton2017value,pont2013measures,sales2017ontological}}

\section*{Acknowledgments}

This work has been partially supported by the Norwegian Research Council grant 329745 Machine Teaching for Explainable AI, also by the EU (FEDER) and Spanish MINECO grant RTI2018-094403-B-C32 funded by  MCIN/AEI/10.13039/501100011033 and by “ERDF A way of making Europe”,  Generalitat Valenciana under grant PROMETEO/2019/098, EU's Horizon 2020 research and innovation programme under grant agreement No. 952215 (TAILOR), US DARPA HR00112120007 (RECoG-AI), and the UPV (Vicerrectorado de Investigación) grant PAI-10-21

\bibliographystyle{unsrtnat}
\bibliography{biblio}

\newpage

\appendix

\section{Supplementary materials}

%%%%%%%%%%%%%%%%%%%%%%%%%%%%%%%%%%%%%%%%%%%%%%%%%%%%%%%%%%%%%%%%%%%%%%%%%%%%%%%%%%%%%%%%%%
%%%%%%%%%%%%%%%%%%%%%%%%%%%%%%%%%%%%%%%%%%%%%%%%%%%%%%%%%%%%%%%%%%%%%%%%%%%%%%%%%%%%%%%%%%
\subsection{Case study 1: Alpha* series}
%%%%%%%%%%%%%%%%%%%%%%%%%%%%%%%%%%%%%%%%%%%%%%%%%%%%%%%%%%%%%%%%%%%%%%%%%%%%%%%%%%%%%%%%%%
%%%%%%%%%%%%%%%%%%%%%%%%%%%%%%%%%%%%%%%%%%%%%%%%%%%%%%%%%%%%%%%%%%%%%%%%%%%%%%%%%%%%%%%%%%
% It is clear that most of the computational effort for the development of AlphaGo Zero and the Alpha Zero was used for generating the training data (self-play games). This data is unclear from AlphaGo Zero paper according to my notes (which I updated after disccusing with Miles et al. a couple of months ago):

% I'm not sure that OpenAI’s figures for Alpha* consider the comp. effort for generating this games (https://blog.openai.com/ai-and-compute/#appendixmethods):  although they estimated some missing figures (operations/FLOPS when GPU time is given ), it isn't the case when neither the number of GPUs nor the time or number of operations performed are provided in the papers (which is the case of AlphaGo Zero).

% So, I think we should mention all of that in the new version of the paper and change the figures considering (or not) the comp. power needed to generate the training games...

\insertedTIST{In this section, we showcase the resources used for the development of the Alpha* systems to play board games. %We analyze the whole series, from  {\sffamily{\small AlphaGo}} \cite{silver2016mastering} (including  the {\sffamily{\small Fan}} and  {\sffamily{\small Lee}} versions) {\sffamily{\small AlphaGo Master}}, {\sffamily{\small AlphaGo Zero}} \cite{silver2017mastering}  and {\sffamily{\small AlphaZero}} \cite{silver2017masteringB}. 
Table \ref{tab:AlphaComp} shows the computational resources for the different Alpha* systems according to the information disclosed in the original papers. It should be noted that most of the computational effort for the development of {\sffamily{\small AlphaGo Master}}, {\sffamily{\small AlphaGo Zero}} and the {\sffamily{\small Alpha Zero}} was used for generating the training data (self-play games). However this data is not explicitly specify %clear\sidenoteJose{Data unclear? Why?} 
for all the systems with the exception of{\sffamily{\small Alpha Zero}}. Therefore,  the compute power used for generating this training data has not been considered for generating the multidimensional utility space for Alpha*.}

\newcolumntype{L}[1]{>{\raggedright\arraybackslash}p{#1}} 
\newcolumntype{C}[1]{>{\centering\arraybackslash}p{#1}}
\newcolumntype{R}[1]{>{\raggedleft\arraybackslash}p{#1}}

\setlength{\tabcolsep}{0.2pt}
\renewcommand{\arraystretch}{0.85}

\begin{table*}[!h]
\begin{threeparttable}[t]
\centering
\resizebox{1\textwidth}{!}{%
\begin{tabular}{L{3.4cm}C{2cm}C{2cm}C{2.5cm}C{2cm}C{3cm}C{1.7cm}cC{2cm}}

\toprule
  \textbf{} & \textbf{Self-Play HW} & \textbf{Self-Play HW (CPU)} & \textbf{Training HW} & \textbf{Training HW (CPU)}  & \textbf{Playing HW} & \textbf{Playing HW (CPU)} & \textbf{ELO rating} & \textbf{Publication Date} \\ 
  \hline 
 \textbf{AlphaGo Fan} \cite{silver2016mastering} & - & -  & 50 GPU & 250 & 176 GPU + 1202 CPU & 2082 & 3144 & Jan 2016 \\ \hline
 \textbf{AlphaGo Lee} \cite{silver2016mastering}& - & - & 50 GPU & 250 & 1920 CPU + 280 GPU & 3320 & 3739 & March 2016 \\ \hline
 \textbf{AlphaGo Master} \cite{silver2016mastering}& - & - &64 GPU & 320 & 4 $TPU_{v1}$ & 240 & 4848 & Dec 2016 \\ \hline 
 \textbf{AlphaGo Zero} \cite{silver2017mastering}& - & - &64 GPU + 19 CPU& 339 & 4 $TPU_{v1}$ & 240 & 5185 & Oct 2017 \\ \hline
 \textbf{Alpha Zero}$\;\;$ \cite{silver2017masteringB}& 5000 $TPU_{v1}$ & 300000 & 64 $TPU_{v2}$ & 11520 & 4 $TPU_{v2}$ & 720 & 5000 & Dec 2017 \\ \toprule
 
\end{tabular}
}
% \begin{tablenotes}
     %\item[1] AlphaGo's policy network was trained on 50 GPUs for three weeks. It was improved by reinforcement learning through self play on 50 GPUs for one day. Its value network was trained on 50 GPUs for one week. Overall that's 4 GPU-years total.
     %\item[2] Each neural network was trained using 64 GPUs and 19 CPUs. The compute power used for executing the self-play in unclear from the paper.
%   \end{tablenotes}
\caption{Computational resources for Alpha* systems. Normalized hardware (in CPUs) has been calculated using the equivalence: $1\;TPU_{v2} \simeq 3\;TPU_{v1} \simeq 36\;GPU \simeq 180\;CPU$  from \cite{jouppi2017datacenter}) } 
\label{tab:AlphaComp}. 
    \end{threeparttable}%
\end{table*}

%%%%%%%%%%%%%%%%%%%%%%%%%%%%%%%%%%%%%%%%%%%%%%%%%%%%%%%%%%%%%%%%%%%%%%%%%%%%%%%%%%%%%%%%%%
%%%%%%%%%%%%%%%%%%%%%%%%%%%%%%%%%%%%%%%%%%%%%%%%%%%%%%%%%%%%%%%%%%%%%%%%%%%%%%%%%%%%%%%%%%
\subsection{Case study 2: ALE papers}
%%%%%%%%%%%%%%%%%%%%%%%%%%%%%%%%%%%%%%%%%%%%%%%%%%%%%%%%%%%%%%%%%%%%%%%%%%%%%%%%%%%%%%%%%%
%%%%%%%%%%%%%%%%%%%%%%%%%%%%%%%%%%%%%%%%%%%%%%%%%%%%%%%%%%%%%%%%%%%%%%%%%%%%%%%%%%%%%%%%%%

\insertedTIST{In this section we examine the resources used for producing the RL-based systems to play the collection of Atari games in ALE benchmark. %We selected all the papers (systems) from EFF's AI Progress Measurement Project \cite{eff2017} and the papers introducing the  {\sffamily{\small Rainbow}} \cite{Rainbow17} and {\sffamily{\small REACTOR}} agents \cite{Reactor17}. 
Table \ref{tab:EFFComp} shows the computational resources for the different systems addressing ALE benchmark according to the information reported in the original papers. In this case, most of the computational effort for the development of the systems fall within the training procedures (CPUs per day and millions of frames used to learn). Whenever the authors do not provide this information explicitly, (e.g., training  time or harware used), this is roughly estimated from the kind of approach used, whether it is follow-up work, the training setting used, etc., or from figures in more recent papers, which make explicit comparisons between them and the state of the art (such as in \cite{Rainbow17,Reactor17}). }

\begin{table*}[!h]
\begin{threeparttable}

  \centering
  \resizebox{1\textwidth}{!}{%
    \begin{tabular}{lC{1.3cm}C{1.5cm}C{1.5cm}C{1.5cm}C{2cm}C{1.4cm}C{1.3cm}C{1.7cm}C{1.8cm}C{2cm}}\toprule
    \textbf{Algorithm} & \textbf{Training Time (days)} & \textbf{Training HW}  & \textbf{Training HW (CPU/days)} &\#\textbf{Workers} & \textbf{Training Frames (Millions)} & \textbf{Games Tested }& \textbf{Frames Test} & \textbf{Test Procedure} & \textbf{Score norm. (median)} & \textbf{Publication Date} \\ \hline
    \textbf{Best linear} \cite{bellemare2013arcade} & 17    & CPU   & 17      & 1     & 0.15  & 50    & 18000 & -  & 14.04 & 19/07/2012 \\\hline
    \textbf{DQN best} \cite{mnih2013playing} & 8     & GPU   & 40       & 1     & 50  & 7     & -     & - & 67.96 & 19/12/2013 \\\hline
    \textbf{Nature DQN} \cite{mnih2015human} & 8     & GPU   & 40      & 1     & 50  & 49    & 18000 & noop  & 92.67 & 26/02/2015 \\\hline
    \textbf{Gorila} \cite{nair2015massively} & 4     & GPU   & 20     & 100   & 200  & 49    & 108000 & hs    & 68.56 & 15/07/2015 \\\hline
    \textbf{DDQN (tuned) hs} \cite{van2016deep} & 8     & GPU   & 40     & 1     & 200  & 57    & 108000 & hs    & 121.04 & 22/09/2015 \\\hline
    \textbf{DQN hs} \cite{van2016deep}& 8     & GPU   & 40     & 1     & 200  & 49    & 108000 & hs    & 68.52 & 22/09/2015 \\\hline
    \textbf{DQN noop} \cite{van2016deep}& 8     & GPU   & 40    & 1     & 200  & 49    & 18000 & noop  & 89.30 & 22/09/2015 \\\hline
    \textbf{Prior hs} \cite{schaul2015prioritized} & 8     & GPU   & 40    & 1     & 200  & 49    & 108000 & hs    & 119.74 & 18/11/2015 \\\hline
    \textbf{Prior noop }\cite{schaul2015prioritized} & 8     & GPU   & 40      & 1     & 200  & 49    & 108000 & noop  & 137.66 & 18/11/2015 \\\hline
    \textbf{DDQN (tuned) noop} \cite{wang2015dueling} & 8     & GPU   & 40     & 1     & 200  & 57    & 18000 & noop  & 132.52 & 20/11/2015 \\\hline
    \textbf{Duel hs} \cite{wang2015dueling} & 8     & GPU   & 40     & 1     & 200  & 57    & 108000 & hs    & 131.66 & 20/11/2015 \\\hline
    \textbf{Duel noop} \cite{wang2015dueling}  & 8     & GPU   & 40      & 1     & 200  & 57    & 18000 & noop  & 166.38 & 20/11/2015 \\\hline
    \textbf{Prior+Duel hs} \cite{wang2015dueling}& 8     & GPU   & 40     & 1     & 200  & 57    & 108000 & hs    & 123.48 & 20/11/2015 \\\hline
    \textbf{Prior+Duel noop} \cite{wang2015dueling} & 8     & GPU   & 40     & 1     & 200  & 57    & 18000 & noop  & 159.73 & 20/11/2015 \\\hline
    \textbf{A3C FF (1 day) hs }\cite{mnih2016asynchronous} & 1     & CPU   & 1     & 16    & 320  & 57    & 108000 & hs    & 78.53 & 04/02/2016 \\\hline
    \textbf{A3C FF hs} \cite{mnih2016asynchronous} & 4     & CPU   & 4    & 16    & 320  & 57    & 108000 & hs    & 117.44 & 04/02/2016 \\\hline
    \textbf{A3C LSTM hs} \cite{mnih2016asynchronous} & 4     & CPU   & 4    & 16    & 320  & 57    & 108000 & hs    & 112.63 & 04/02/2016 \\\hline
    \textbf{DDQN+Pop-Art noop }\cite{van2016learning} & 8     & GPU   & 40    & 1     & 200  & 49    & 108000 & noop  & 112.83 & 24/02/2016 \\\hline
    \textbf{A3C-CTS} \cite{bellemare2016unifying} & 4     & CPU   & 4    & 16    & 200  & 60    & -     & - & 89.67 & 06/06/2016 \\\hline
    \textbf{SARSA} \cite{bellemare2012investigating} & 30    & CPU   & 30    & 1     & 2   & 46    & -     & -  & 10.09 & 06/06/2016 \\\hline
    \textbf{TRPO-hash }\cite{tang2017exploration} & 8     & GPU   & 40     & 1     & 50  & 6     & -     & - & 28.90 & 15/11/2016 \\\hline
    \textbf{DQN-CTS} \cite{ostrovski2017count} & 8     & GPU   & 40    & 1     & 150  & 57    & -     & - & 11.98 & 03/03/2017 \\\hline
    \textbf{DQN-PixelCNN} \cite{ostrovski2017count} & 8     & GPU   & 40   & 1     & 150  & 57    & -     & - & 27.45 & 03/03/2017 \\\hline
    \textbf{ES FF (1 hour) noop} \cite{salimans2017evolution} & 0.0416 & CPU   & 0.0416  &   1    & 1000  M & 51    & 108000 & noop  & 40.18 & 10/03/2017 \\\hline
    \textbf{REACTOR hs} \cite{gruslys2018reactor} & 2     & CPU   & 2    & 10 + 1\tnote{1} & 200  & 57    & 108000 & hs    & 154.42 & 15/04/2017 \\\hline
    \textbf{REACTOR 500M hs} \cite{gruslys2018reactor}& 4     & CPU   & 4     & 10 + 1\tnote{1} & 500  & 57    & 108000 & hs    & 185.56 & 15/04/2017 \\\hline
    \textbf{REACTOR noop} \cite{gruslys2018reactor}& 2     & CPU   & 2    & 10 + 1\tnote{1} & 200  & 57    & 18000 & noop  & 185.95 & 15/04/2017 \\\hline
    \textbf{REACTOR 500M noop} \cite{gruslys2018reactor}& 4     & CPU   & 4     & 10 + 1\tnote{1} & 500  M & 57    & 18000 & noop  & 291.74 & 15/04/2017 \\\hline
    \textbf{Sarsa-e} \cite{martin2017count} & 8     & GPU   & 40   & 1     & 100  M & 5     & 18000 & noop  & 28.07 & 25/06/2017 \\\hline
    \textbf{Sarsa-f-EB} \cite{martin2017count} & 8     & GPU   & 40     & 1     & 100  M & 5     & 18000 & noop  & 62.86 & 25/06/2017 \\\hline
    \textbf{C51 noop} \cite{bellemare2017distributional} & 8     & GPU   & 40     & 1     & 200  M & 57    & 18000 & noop  & 177.71 & 21/07/2017 \\\hline
    \textbf{Rainbow hs} \cite{hessel2017rainbow} & 10    & GPU   & 40     & 1     & 200  M & 54    & 108000 & hs    & 144.96 & 06/10/2017 \\\hline
    \textbf{Rainbow noop} \cite{hessel2017rainbow} & 10    & GPU   & 40    & 1     & 200  M & 54    & 18000 & noop  & 232.46 & 06/10/2017 \\\hline
    \toprule
    \end{tabular}%
    }
\begin{tablenotes}
     \item[1] \footnotesize{10 actor-learner workers (CPUs) and 1 parameter server.}
     %\item[2] Each neural network was trained using 64 GPUs and 19 CPUs. The compute power used for executing the self-play in unclear from the paper.
\end{tablenotes}
  \caption{Computational resources for the systems addressing ALE benchmark \cite{bellemare13arcade}. Systems from \cite{eff2017,Rainbow17,Reactor17}. Normalized training time (in CPUs) has been calculated using the equivalence: $1\; GPU \simeq 5\; CPU$ from \cite{jouppi2017datacenter}. \emph{\#Workers} represents the number of parallel machines used. We calculated median human normalised scores across all games according to \cite{nair2015massively}. }
  \label{tab:EFFComp}%
\end{threeparttable}%
\end{table*}%

%%%%%%%%%%%%%%%%%%%%%%%%%%%%%%%%%%%%%%%%%%%%%%%%%%%%%%%%%%%%%%%%%%%%%%%%%%%%%%%%%%%%%%%%%%
%%%%%%%%%%%%%%%%%%%%%%%%%%%%%%%%%%%%%%%%%%%%%%%%%%%%%%%%%%%%%%%%%%%%%%%%%%%%%%%%%%%%%%%%%%
\subsection{Case study 3: ImageNet papers}
%%%%%%%%%%%%%%%%%%%%%%%%%%%%%%%%%%%%%%%%%%%%%%%%%%%%%%%%%%%%%%%%%%%%%%%%%%%%%%%%%%%%%%%%%%
%%%%%%%%%%%%%%%%%%%%%%%%%%%%%%%%%%%%%%%%%%%%%%%%%%%%%%%%%%%%%%%%%%%%%%%%%%%%%%%%%%%%%%%%%%

\insertedTIST{In this section we disclose the resources used by systems addressing the third case study: the Imagenet Large Scale Visual Recognition Challenge (ILSVRC). %, for which we analyse We analysed the following DNNs which obtained the highest performance in the  past six years of the challenge, including those more recent approaches developed for environments with and small computational budget (e.g., mobile devices): {\sffamily{\small AlexNet}} \cite{krizhevsky2012imagenet}, {\sffamily{\small BN-AlexNet}} \cite{Zagoruyko2016}, {\sffamily{\small BN-NiN}} \cite{lin2013network}, {\sffamily{\small ENet}} \cite{paszke2016enet}, {\sffamily{\small GoogLeNet}} \cite{szegedy2015going}, {\sffamily{\small VGG}} \cite{simonyan2014very}, {\sffamily{\small ResNet}} \cite{he2016deep}, {\sffamily{\small Inception-v3}} \cite{szegedy2016rethinking}, {\sffamily{\small Inception-v4}} \cite{szegedy2017inception}, {\sffamily{\small Shufflenet}} \cite{zhang1707shufflenet}, {\sffamily{\small Mobilenet-v1}} \cite{howard2017mobilenets}, {\sffamily{\small Mobilenet-v2}} \cite{sandler2018mobilenetv2}, {\sffamily{\small Xception}} \cite{chollet2017xception}, {\sffamily{\small Densenet}} \cite{huang2017densely}, {\sffamily{\small Squeezenet}} \cite{iandola2016squeezenet}, {\sffamily{\small fd-MobileNet}} \cite{qin2018fd}, {\sffamily{\small AmoebaNet}} \cite{real2018regularized}, {\sffamily{\small SENet}} \cite{hu2018squeeze}, {\sffamily{\small Shufflenet v2}} \cite{ma2018shufflenet}, {\sffamily{\small GPipe}} \cite{huang2018gpipe} and {\sffamily{\small PolyNet}} \cite{zhang2017polynet} 
Table \ref{tab:ImageNetComp} shows a number of metrics related to the computational resources used by the different approaches (training hardware, number of layers, number of training images, training time  and number of operations) as well as their load/size (number of parameters learned) and the energy consumption, according to the information disclosed in the original papers or in different sources (possibly from different authors).}

\begin{table*}[!h]
\begin{threeparttable}[t]
\centering
\resizebox{1\textwidth}{!}{%
%\begin{tabular}{C{2.8cm}C{2cm}C{2cm}C{2cm}C{2cm}C{1.7cm}C{1.7cm}cC{2cm}}
\begin{tabular}{lC{3.8cm}C{1.5cm}C{2cm}C{1cm}C{2cm}lC{2cm}C{1cm}c}

\toprule

\textbf{System} & \textbf{Training HW} & \textbf{\#Layers} & \textbf{\#Params (Millions)} & \textbf{Batch Size} & \textbf{Training time} & \textbf{G-Ops$^*$} & \textbf{Net power (Watts)$^*$} & \textbf{Top-1 ACC$^*$} & \textbf{Date} \\ 
  \hline

\textbf{AlexNet} \cite{krizhevsky2012imagenet} & 2 $\times$ nVidia GTX580  &   8 & 60 & 128 & 6 days & 2.26 & 11.20 & 54.61 & 03/12/2012 \\ \hline
\textbf{BN-NIN} \cite{lin2013network} & - &   8 & 8.6 & 128 & - & 2.52 & 12.60 & 62.62 & 16/12/2013 \\ \hline
\textbf{VGG-16} \cite{simonyan2014very}& 4 $\times$ nVidia Titan Black &  16 & 138.4 &  64 & 2-3 weeks & 30.97 & 12.40 & 70.62 & 04/9/2014 \\ \hline
\textbf{VGG-19} \cite{simonyan2014very}& 4 $\times$ nVidia Titan Black &  19 & 143.7 & 128 & 2-3 weeks & 39.29 & 12.20 & 70.74 & 04/9/2014 \\ \hline
\textbf{GoogLeNet} \cite{szegedy2015going} & few GPUs &  22 & 7 & 128 & 1 week & 3.00 & 11.15 & 68.70 & 17/9/2014 \\ \hline
\textbf{BN-AlexNet} \cite{Zagoruyko2016} & - &   8 & 60.6 & 256 & - & 1.30 & 11.40 & 56.60 & 11/2/2015 \\ \hline
\textbf{Inception-v3} \cite{szegedy2016rethinking}& 50 $\times$ NVidia Kepler &  42 & 23.85 & 256 & - & 11.45 & 12.30 & 78.53 & 02/12/2015 \\ \hline
\textbf{ResNet-18} \cite{he2016deep}& few GPUs &  18 & 11.7 & 256 & - & 3.63 & 12.30 & 69.57 & 10/12/2015 \\ \hline
\textbf{ResNet-34} \cite{he2016deep}& few GPUs &  34 & 21.8 & 256 & - & 7.34 & 12.80 & 73.27 & 10/12/2015 \\ \hline
\textbf{ResNet-50} \cite{he2016deep}& few GPUs &  50 & 25.6 & 256 & - & 8.21 & 11.80 & 75.99 & 10/12/2015 \\ \hline
\textbf{ResNet-101} \cite{he2016deep}& few GPUs & 101 & 44.6 & 256 & - & 15.65 & 11.45 & 77.56 & 10/12/2015 \\ \hline
\textbf{ResNet-152} \cite{he2016deep}& few GPUs & 152 & 60.3 & 256 & - & 23.10 & 11.40 & 77.84 & 10/12/2015 \\ \hline
\textbf{Inception-v4} \cite{szegedy2017inception}& 20 $\times$ NVidia Kepler &  75 & 31.6 & 256 & - & 18.44 & 11.60 & 80.10 & 23/2/2016 \\ \hline
\textbf{ENet} \cite{paszke2016enet} & 4 nVidia Titan X &  29 & 5.9 & 128 & 3- 6  hours & 1.57 & 11.60 & 68.40 & 7/6/2016 \\ \hline
\textbf{DenseNet-121} \cite{huang2017densely}& 8 $\times$ nVidia Tesla M40 & 121 & 8 & 256 & - & 5.71 & - & 74.98 & 25/8/2016 \\ \hline
\textbf{DenseNet-169} \cite{huang2017densely}& 8 $\times$ nVidia Tesla M40 & 169 & 14.2 & 256 & - & 6.77 & - & 76.20 & 25/8/2016 \\ \hline
\textbf{DenseNet-201} \cite{huang2017densely}& 8 $\times$ nVidia Tesla M40 & 201 & 20 & 256 & - & 8.65 & - & 77.42 & 25/8/2016 \\ \hline
\textbf{Xception} \cite{chollet2017xception}& 60 $\times$ nVidia K80 &  36 & 22.9 & 256 & 3 days & 16.75 & - & 79.00 & 07/10/2016 \\ \hline
\textbf{SqueezeNet} \cite{iandola2016squeezenet}& - &   8 & 1.3 & 512 & - & 1.64 & - & 57.50 & 04/11/2016 \\ \hline

\textbf{PolyNet} \cite{zhang2017polynet}& 32 $\times$ nVidia Titan X &  92 & 2.9 & 512 & - & - & - & 81.29 & 17/11/2016 \\  \hline
\textbf{MobileNet-v1} \cite{howard2017mobilenets} & - &  14 & 4.3 &  96 & - & 1.14 & - & 70.90 & 17/4/2017 \\ \hline
\textbf{ShuffleNet} \cite{zhang1707shufflenet} & 4 $\times$ GPUs &   8 & 1.9 & 1024 & 1-2 days & 1.57 & - & 62.20 & 04/7/2017 \\ \hline
\textbf{MobileNet-v2} \cite{sandler2018mobilenetv2}& 16 $\times$ GPU &  20 & 3.5 &  96 & - & 0.60 & - & 71.80 & 13/1/2018 \\ \hline

\textbf{fd-MobileNet} \cite{qin2018fd}& 4 $\times$ GPUs &  12 & 2.9 & 256 & - & 0.29 & - & 65.30 & 11/2/2018 \\ \hline

\textbf{AmoebaNet} \cite{real2018regularized}& 450 $\times$ nVidia K40 &  - & 469 & 256 & 7 days & - & - & 83.9 & 05/02/2018 \\  \hline
\textbf{Shufflenet v2} \cite{ma2018shufflenet}& 64 $\times$ nVidia Titan Pascal &  164 & 137 & 4 & - & - & - & 81.44\tnote{1} & 30/07/2018 \\  \hline

\textbf{SENet} \cite{hu2018squeeze}& 8 $\times$ nVidia Titan X &  154 & 145.8 & 256 & - & - & - & 80.19\tnote{1} & 05/09/2018 \\\hline

\textbf{GPipe} \cite{huang2018gpipe}& TPU$_v2$ &  - & 557 & 256 & - & - & - & 84.3\tnote{1} & 16/11/2018 \\

 \toprule
 
\end{tabular}
}
\begin{tablenotes}
     \item[1] \footnotesize{Results over the ImageNet validation set.}
     %\item[2] Each neural network was trained using 64 GPUs and 19 CPUs. The compute power used for executing the self-play in unclear from the paper.
\end{tablenotes}
\caption{Computational, load, and energy resources for ImageNet systems. Some of the values for the attributes with $*$ were obtained from \cite{canziani2016analysis}.} 
\label{tab:ImageNetComp}. 
    \end{threeparttable}%
\end{table*}

%%%%%%%%%%%%%%%%%%%%%%%%%%%%%%%%%%%%%%%%%%%%%%%%%%%%%%%%%%%%%%%%%%%%%%%%%%%%%%%%%%%%%%%%%%
%%%%%%%%%%%%%%%%%%%%%%%%%%%%%%%%%%%%%%%%%%%%%%%%%%%%%%%%%%%%%%%%%%%%%%%%%%%%%%%%%%%%%%%%%%
\subsection{Case study 4: Personal Assistants}
%%%%%%%%%%%%%%%%%%%%%%%%%%%%%%%%%%%%%%%%%%%%%%%%%%%%%%%%%%%%%%%%%%%%%%%%%%%%%%%%%%%%%%%%%%
%%%%%%%%%%%%%%%%%%%%%%%%%%%%%%%%%%%%%%%%%%%%%%%%%%%%%%%%%%%%%%%%%%%%%%%%%%%%%%%%%%%%%%%%%%

\insertedTIST{In this section we display some of the costs (power requirements in deployment) and performance of those virtual personal assistants analysed in the fourth case study. Table \ref{tab:PAenergyPerf} shows the values of this limited number of resources, mostly collected from external sources.}

\begin{table*}[!h]
\begin{threeparttable}[t]
\centering
\resizebox{1\textwidth}{!}{%
%\begin{tabular}{C{2cm}C{2cm}C{2cm}C{2cm}C{2cm}C{1.7cm}C{1.7cm}cC{2cm}}
\begin{tabular}{lC{2cm}C{2cm}C{2.5cm}C{2.2cm}C{2cm}C{2cm}C{2cm}C{2cm}C{2cm}}

\toprule

\textbf{System} &  \textbf{Power Req. (Watts)} & \textbf{Power Idle (Watts)} & \textbf{Power Playing music (Watts)} & \textbf{Q. Attempted  (2017)} & \textbf{Q. Correct (2017)} & \textbf{Q. Attempted  (2018)} & \textbf{Q. Correct (2018)} & \textbf{Date} \\ 
  \hline
\textbf{Apple Siri}  & - & - & 9.25 & 31.40 & 86.10 & 40.80 & 80.00 & 12/10/2011 \\ \hline
\textbf{Microsoft Cortana}  & - & - & - & 53.90 & 86.00 & 64.90 & 91.60 & 2/4/2014 \\ \hline
\textbf{Amazon 1st-Gen Echo } & 21.00 & 2.95 & 3.25 & 19.80 & 94.50 & 53.70 & 86.20 & 6/11/2014 \\ \hline
\textbf{Amazon 1st-Gen Echo Dot}  & 9.36 & 1.75 & 2.25 & 19.80 & 94.50 & 53.70 & 86.20 & 1/3/2016 \\ \hline
\textbf{Google Assistant}  & - & - & - & - & - & 77.20 & 95.20 & 1/5/2016 \\ \hline
\textbf{Amazon 2nd-Gen Echo Dot} & 9.36 & 1.75 & 2.25 & 19.80 & 94.50 & 53.70 & 86.20 & 20/10/2016 \\ \hline
\textbf{Google Home}  & 33.00 & 2.00 & - & 65.30 & 94.50 & 66.20 & 87.50 & 4/11/2016 \\ \hline
\textbf{Amazon Echo Spot}  & 10.92 & 2.08 & 2.90 & 19.80 & 94.50 & 53.70 & 86.20 & 27/9/2017 \\ \hline
\textbf{Amazon 2nd-Gen Echo} & 21.00 & 1.95 & 2.90 & 19.80 & 94.50 & 53.70 & 86.20 & 1/10/2017 \\ \hline
\textbf{Google Home Mini}  & 9.00 & 1.50 & 2.25 & 65.30 & 94.50 & 66.20 & 87.50 & 4/10/2017 \\ \hline
\textbf{Google Home Max } & - & - & - & 65.30 & 94.50 & 66.20 & 87.50 & 4/10/2017 \\ \hline
\textbf{Amazon Echo Plus}  & 30.00 & 2.40 & 3.65 & 19.80 & 94.50 & 53.70 & 86.20 & 31/10/2017 \\ \hline
\textbf{Apple HomePod}  & - & 1.76 & 9.25 & 31.40 & 86.10 & 40.80 & 80.00 & 9/2/2018 \\ \hline
\textbf{Amazon 3rd-Gen Echo Dot}  & 15.00 & - & - & 19.80 & 94.50 & 53.70 & 86.20 & 1/9/2018 \\ \hline
\textbf{Google Home Hub}  & 33.00 & - & -  & 65.30 & 94.50 & 66.20 & 87.50 & 9/10/2018 \\ \hline

 \toprule
 
\end{tabular}
}
%\begin{tablenotes}
     %\item[1] 
%\end{tablenotes}
\caption{Energy resources for personal assistants. Performance values (questions attempted and questions correct) are from \cite{StoneTemple2018} where the digital marketing firm \textit{Stone Temple}  tested the assistants via an exhaustive list of 4,942 queries. Power requirements have been collected from manuals and other sources. Power consumptions (iddle and playing music) are from \cite{HTG2018,ambient2018}
} 
\label{tab:PAenergyPerf}. 
    \end{threeparttable}%
\end{table*}

\end{document}